\documentclass[pdflatex,sn-mathphys-num]{sn-jnl}

\usepackage{graphicx}%
\usepackage{multirow}%
\usepackage{amsmath,amssymb,amsfonts}%
\usepackage{amsthm}%
\usepackage{mathrsfs}%
\usepackage[title]{appendix}%
\usepackage{xcolor}%
\usepackage{textcomp}%
\usepackage{manyfoot}%
\usepackage{booktabs}%
\usepackage{algorithm}%
\usepackage{algorithmicx}%
\usepackage{algpseudocode}%
\usepackage{listings}%
\usepackage{float}

\usepackage[numbers]{natbib}
\usepackage{caption}
\usepackage{subcaption}

\theoremstyle{thmstyleone}%
\theoremstyle{thmstyletwo}%

\theoremstyle{thmstylethree}%

\begin{document}

\title{BabyFlow: 3D modeling of realistic and expressive infant faces} 


\author*[1]{\fnm{Antonia} \sur{Alomar}}\email{antonia.alomar@upf.edu}

\author[1]{\fnm{Mireia} \sur{Masias}}

\author[2,3]{\fnm{Marius George} \sur{Linguraru}}

\author[1]{\fnm{Federico M.} \sur{Sukno}}

\author[1]{\fnm{Gemma} \sur{Piella}}

\affil*[1]{\orgdiv{Department of Engineering}, \orgname{Universitat Pompeu Fabra}, \orgaddress{\street{122-140 Tànger}, \city{Barcelona}, \postcode{08018},  \country{Spain}}}

\affil[2]{\orgdiv{Sheikh Zayed Institute for Pediatric Surgical Innovation}, \orgname{Children's National Hospital}, \orgaddress{\street{111 Michigan Ave NW}, \city{Washington}, \postcode{20010}, \state{DC}, \country{USA}}}

\affil[3]{\orgdiv{Department of Radiology and Pediatrics}, \orgname{George Washington University}, \orgaddress{\street{2121 I St NW}, \city{Washington}, \postcode{20052}, \state{DC}, \country{USA}}}

\abstract{
Early detection of developmental disorders can be aided by analyzing infant craniofacial morphology, but modeling infant faces is challenging due to limited data and frequent spontaneous expressions. We introduce BabyFlow, a generative AI model that disentangles facial identity and expression, enabling independent control over both. Using normalizing flows, BabyFlow
learns flexible, probabilistic representations that capture the complex, non-linear variability of expressive infant faces without restrictive linear assumptions. To address scarce and uncontrolled expressive data, we perform cross-age expression transfer, adapting expressions from adult 3D scans to enrich infant datasets with realistic and systematic expressive variants. As a result, BabyFlow improves 3D reconstruction accuracy, particularly in highly expressive regions such as the mouth, eyes, and nose, and supports synthesis and modification of infant expressions while preserving identity. Additionally, by integrating with diffusion models, BabyFlow generates high-fidelity 2D infant images with consistent 3D geometry, providing powerful tools for data augmentation and early facial analysis.
}




\maketitle

\section{Introduction}\label{sec1}

Craniofacial features play a critical role in early detection of developmental disorders. Recent studies have shown that subtle differences in facial structure, known as craniofacial dysmorphology, can serve as early indicators of genetic syndromes and other developmental disturbances \cite{LearnedMiller2006, Tu2018, Waddington2025, Hallgrimsson2020}. Identifying these differences as early as possible is essential to better support the medical, functional, and developmental needs of affected children \cite{Huang2016, Collett2021, Junaid2022}. For this reason, there is a growing focus on developing diagnostic tools that can be used during the perinatal stages of life.

One promising direction involves analyzing the 3D shape of infant faces using imaging technologies. However, capturing suitable 3D scans of infant faces for analysis is particularly challenging. Studies of facial features typically rely on a neutral facial expression (i.e., free of smiles, frowns, or other emotions) to ensure consistency and comparability across individuals \cite{LearnedMiller2006, Waddington2025, Hammond2012, Matthews2021}. Yet, obtaining such expressionless faces in newborns and infants is difficult due to their limited cooperation and frequent spontaneous facial movements. This, together with severe data scarcity, makes it hard to build reliable models of infant face geometry, limiting their use in clinical settings \cite{Hermann2016, Brons2019}. 

At the same time, infant facial expressions are themselves clinically valuable. Early facial expressivity has been linked to neurodevelopment and can serve as a proxy for assessing brain and central nervous system maturation \cite{Kanenishi2013, Lin2020, book_emotion_recognition}. This dual importance underscores the need for an expressive and controllable infant face model that can separate expression from identity. Such a model, capable of modifying facial expression while preserving identity, would enable more precise and comprehensive analyses of both facial morphology and expression. Importantly, it could be used to neutralize non-neutral scans, reducing data exclusion and avoiding the need to wait for a naturally neutral expression during acquisition. This would improve data usability and efficiency in clinical and research settings.

Building such an expressive and controllable model for face analysis requires a solid foundation: a reference model that faithfully captures the typical craniofacial anatomy of the infant population. To support this type of analysis, 3D Morphable Models (3DMMs) are widely used to represent (adult) facial shape variation in a compact and structured way \cite{Blanz1999, MoralesCompSciRev2021}. 3DMMs learn a mapping of facial shapes into a continuous latent space that also allows synthesizing new realistic facial shapes by reversing the learnt mapping. However, for these models to be useful, they must be built using data that closely match the characteristics of the target group, especially age, but also sex and ethnicity. Therefore, models built from adult faces perform poorly when applied to infants \cite{Sariyanidi2022, Morales2025}. This is a serious limitation, especially when trying to detect subtle facial differences related to developmental conditions, where accuracy is critical. To address this limitation, we have recently developed BabyFM \cite{Morales2025}, the first 3DMM built exclusively from infant face scans. BabyFM effectively models normal facial morphology in infants, enabling more accurate analysis than adult-based models. However, BabyFM does not capture facial expressivity and lacks the capacity to manipulate identity and expression independently. 

To overcome the above limitations, in this paper we present BabyFlow, a new artificial intelligence (AI) model designed to represent expressive infant faces with independent control over identity and expression. This requires moving beyond the classical formulation of 3DMMs, whose linear mapping into a unique subspace, typically assumed to be Gaussian, is insufficient to represent the diversity and complexity of real expressive faces \cite{Egger2020, MoralesCompSciRev2021, Bannister2022, Dnkel2024NormalizingFO}. 

Recent advances in AI, particularly deep learning, offer more flexible and expressive generative modeling approaches. Methods such as variational autoencoders (VAEs), generative adversarial networks (GANs), and diffusion models have shown remarkable promise in producing realistic facial images \cite{Sharma2022, Lattas2023, Ling2023, Stypulkowski_2024}. However, these models either approximate the data distribution (as in VAEs), do not model it explicitly (as in GANs), or involve computationally intensive inference for likelihood estimation (as in diffusion models). As a result, they offer limited control over the underlying density function, making it challenging to evaluate the likelihood of a given face or to manipulate specific aspects of facial variation in a principled way. In contrast, normalizing flows (NFs) are a class of deep learning generative models that are invertible by design, provide exact likelihood estimation, efficient sampling, and allow direct manipulation of the data distribution \cite{Kobyzev2021}. These properties make NFs especially well suited for modeling the fine-grained variations in infant facial morphology and expression.

Building on the strengths of NFs, we present BabyFlow, an expressive and controllable AI model of infant facial shape and expression. Our key contributions are:

\begin{itemize}
    \item Improved 3D representation of infant facial expressivity: By explicitly modeling identity and expressions in separate subspaces, BabyFlow achieves lower reconstruction errors compared to state-of-the-art models, particularly in areas related to facial expressions such as the mouth, eyes, and nose.

    \item Cross-age expression transfer for structured data augmentation: Factoring identity and expression variations enables the synthesis of expressive 3D infant faces, allowing arbitrary modification of expressions while preserving identity, or vice versa, and enriching datasets with realistic variability.
  
    \item Enhanced control for synthetic infant 2D image generation: When combined with diffusion models, BabyFlow allows creating realistic, high-fidelity 2D infant images with known 3D geometry under arbitrary expressions, providing a valuable tool for AI training without the need for collecting massive datasets.

    \item Probabilistic modeling of infant face geometry with normalizing flows: NFs learn to transform the arbitrarily complex subspaces of identity and expressions into simple Gaussian distributions, producing well-behaved latent representations that facilitate sampling and other mathematical operations, such as interpolation.
    
\end{itemize}
Together, these innovations form a flexible framework for infant craniofacial analysis, enabling accurate face reconstruction and realistic synthesis while addressing challenges related to data scarcity, expression modeling, and the infant-specific nature of the task. 
\begin{figure}[ht]
  \centering
        \includegraphics[width= 1 \textwidth]{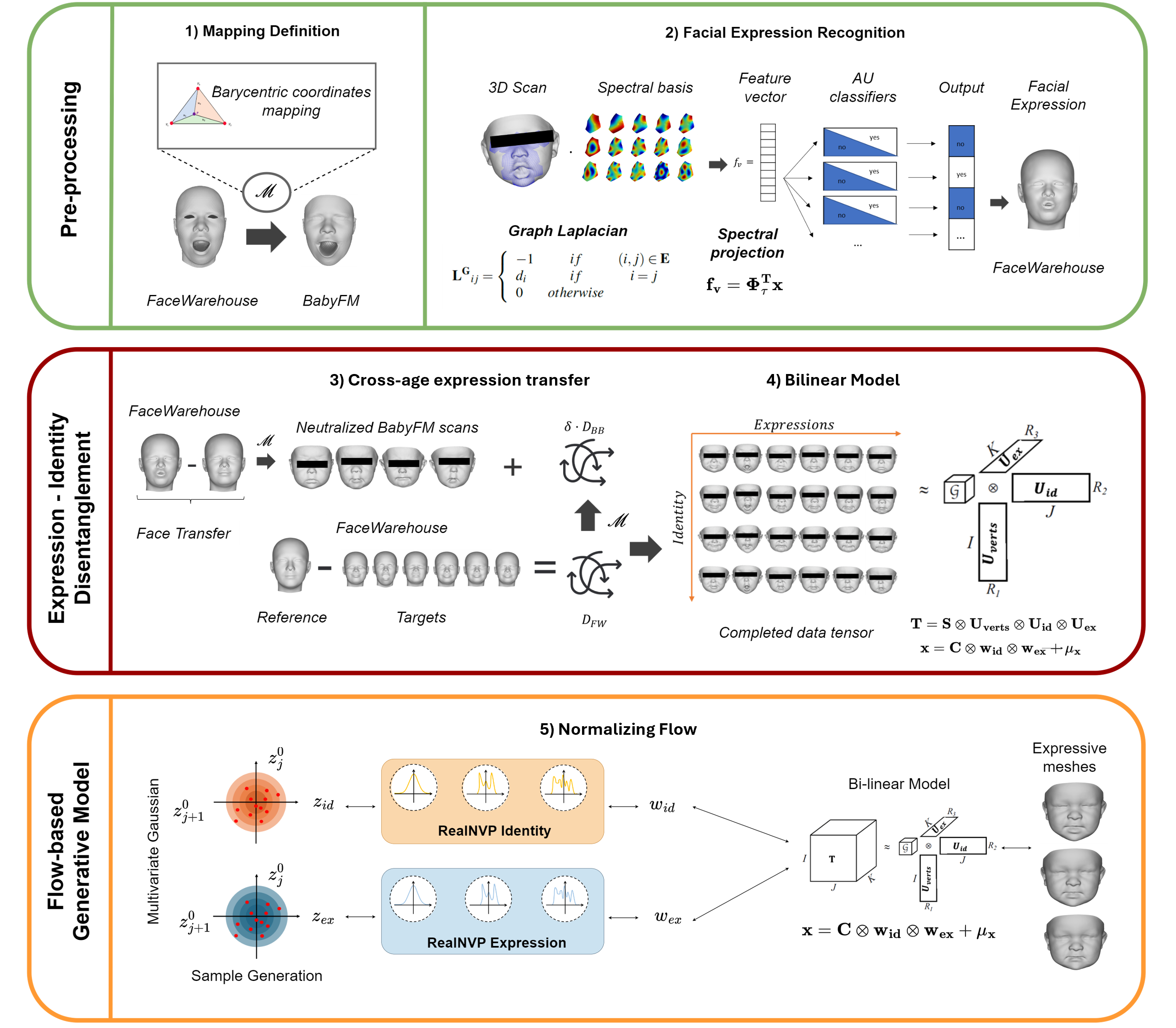}
       \caption{
       Overview of the BabyFlow's pipeline: 1) estimate a mapping $ \mathbf{\mathscr{M}}$ that uses barycentric coordinates to align the adult FaceWarehouse \cite{Zhou2014} meshes with the BabyFM \cite{Morales2025} mesh topology; 2) detect facial expressions by recognizing action units (AUs) \cite{Ekman2002} using a spectral representation of 3D geometry; 3) transfer a full set of expressions from the adult surfaces to the infant meshes; 4) apply a tensor factorization technique, to separate identity and expression components, resulting in a bilinear model; 5) use normalizing flows to model and sample from the latent spaces of identity and expression, enabling the generation of realistic and expressive infant faces.}
      \label{fig:BabyX}
\end{figure} 

\section{Results}\label{sec:results}
The following subsections present the results obtained from the proposed methodology described in Section \ref{sec:meth_data}-\ref{sec:meth_NF} and the experimental setup detailed in Section \ref{sec:experiments}. We evaluated the performance of our approach through both quantitative and qualitative analyses, focusing on reconstruction and generative capabilities, identity-expression disentanglement, and latent space behavior. 
\subsection{BabyFlow: a flow-based model for expressive infant faces}

We present BabyFlow, a deep generative model specifically designed to generate and manipulate expressive infant faces with independent control over identity and expression. The pipeline is illustrated in Figure \ref{fig:BabyX}. The process begins by mapping the mesh topology between adult and infant scans (Step 1) to establish a common mesh structure, allowing transfer learning from the well-defined and largely available adult expressions and data. Next, infant facial expressions are automatically detected (Step 2) to determine each face’s baseline state, after which adult expressions are transferred to generate a complete identity–expression tensor (Step 3). This tensor is subsequently factorized into separate identity and expression components using a bilinear decomposition \cite{Vlasic2005}, resulting in a infant-specific 3DMM that supports facial expression manipulation (Step 4). To enhance generative capabilities, both identity and expression components are modeled using NFs (Step 5), which transform the complex and unknown data distributions into Gaussian latent spaces, overcoming the limitations of traditional 3DMMs in representing non-linear and expressive facial variations. The result is BabyFlow, a powerful, expressive, and controllable flow-based AI model of infant faces.
 
\subsection{BabyFlow can perform cross-age expression transfer from adult-to-infant}
\begin{figure}[!ht]
  \centering
        \includegraphics[width= 0.95\textwidth]{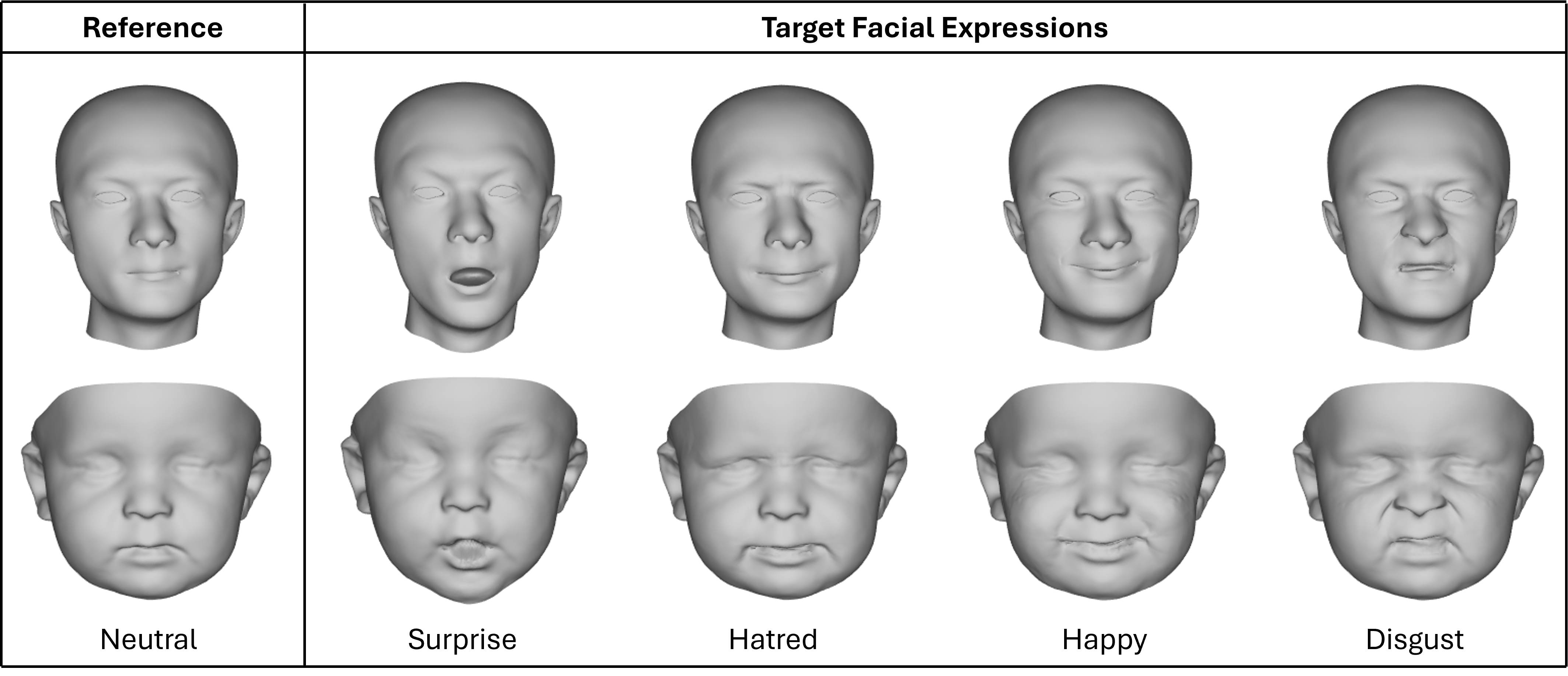}
       \caption{Cross-age expression transfer from adults to infants. The first column shows a baby an infant identity (bottom row) and an adult identity (top row), both displaying a neutral expression.
        In the remaining columns, the top row presents various target expressions of the adult subject, and the bottom row shows the corresponding expressions transferred from the adult to the infant.}
      \label{fig:cross_age_transfer}
\end{figure} 
To evaluate BabyFlow’s ability to transfer facial expressions from adults to infants, Figure~\ref{fig:cross_age_transfer} presents the results of transferring expressions such as surprise and disgust from the FaceWarehouse (FW) model \cite{Zhou2014} to infant faces. The top row shows the target adult facial expressions, while the bottom row displays the corresponding expressions transferred to the infant identity. The results demonstrate that the model successfully transfers the intended facial expressions from adults to infants without introducing artifacts, distortions, or unrealistic facial deformations. 

\subsection{BabyFlow reduces 3D reconstruction errors in key expressive areas} \label{sec:results_3Drec}
We first evaluated BabyFlow’s ability to reconstruct 3D facial geometry from unseen infant scans and compared its performance against the current state-of-the-art model, BabyFM \cite{Morales2025}. The 3D reconstruction process involves fitting the models (BabyFlow and BabyFM) to the test set (20 3D infant scans not used during training) by optimizing the latent parameters that best represent the facial geometry of each subject. Quantitative evaluation was performed by computing the per-vertex reconstruction error between the original 3D scans and the reconstructions generated by each model. 

\begin{figure}[ht]
    \centering
    \includegraphics[width=0.85\textwidth]{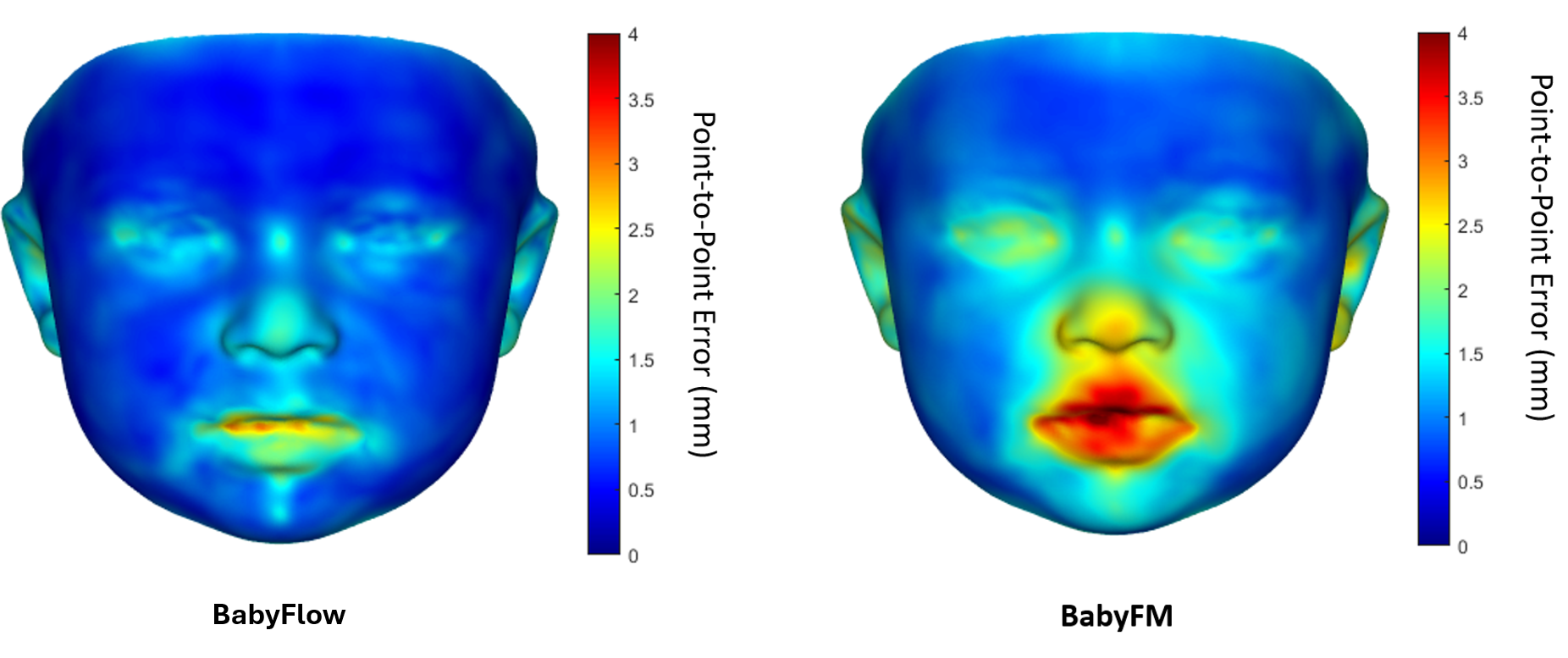}
    \caption{Mean error per vertex across the test set, obtained by fitting the 3D scans using the BabyFlow (left) and the BabyFM \cite{Morales2025} (right). Warmer colors correspond to higher errors.}  \label{fig:error_reconstruction}
    \end{figure}

Figure \ref{fig:error_reconstruction} presents the mean per-vertex reconstruction error across the test set. BabyFlow achieved a mean reconstruction error of 0.928 $\pm$  0.142 mm, significantly outperforming BabyFM, which yielded an error of 1.567 $\pm$  0.384 mm. The highest errors (colored in red) for both models are observed in the mouth region, likely due to its inherent expressiveness and the presence of noise in the 3D scans. However, BabyFlow consistently achieves lower errors, particularly in regions such as the mouth and nose, which are more susceptible to expression-induced deformations. In contrast, BabyFM, trained exclusively on scans with uncontrolled expressions, struggles to accurately reconstruct these more dynamic facial areas.

These results highlight BabyFlow’s superior generalization to unseen subjects and its enhanced capacity to capture expression-induced facial variations, leading to more accurate and realistic 3D face reconstructions.

\subsection{BabyFlow synthesizes realistic, expressive, and controllable 3D infant faces} \label{sec:results_synth}

\begin{figure}[ht]
  \centering
        \includegraphics[width= 0.95 \textwidth]{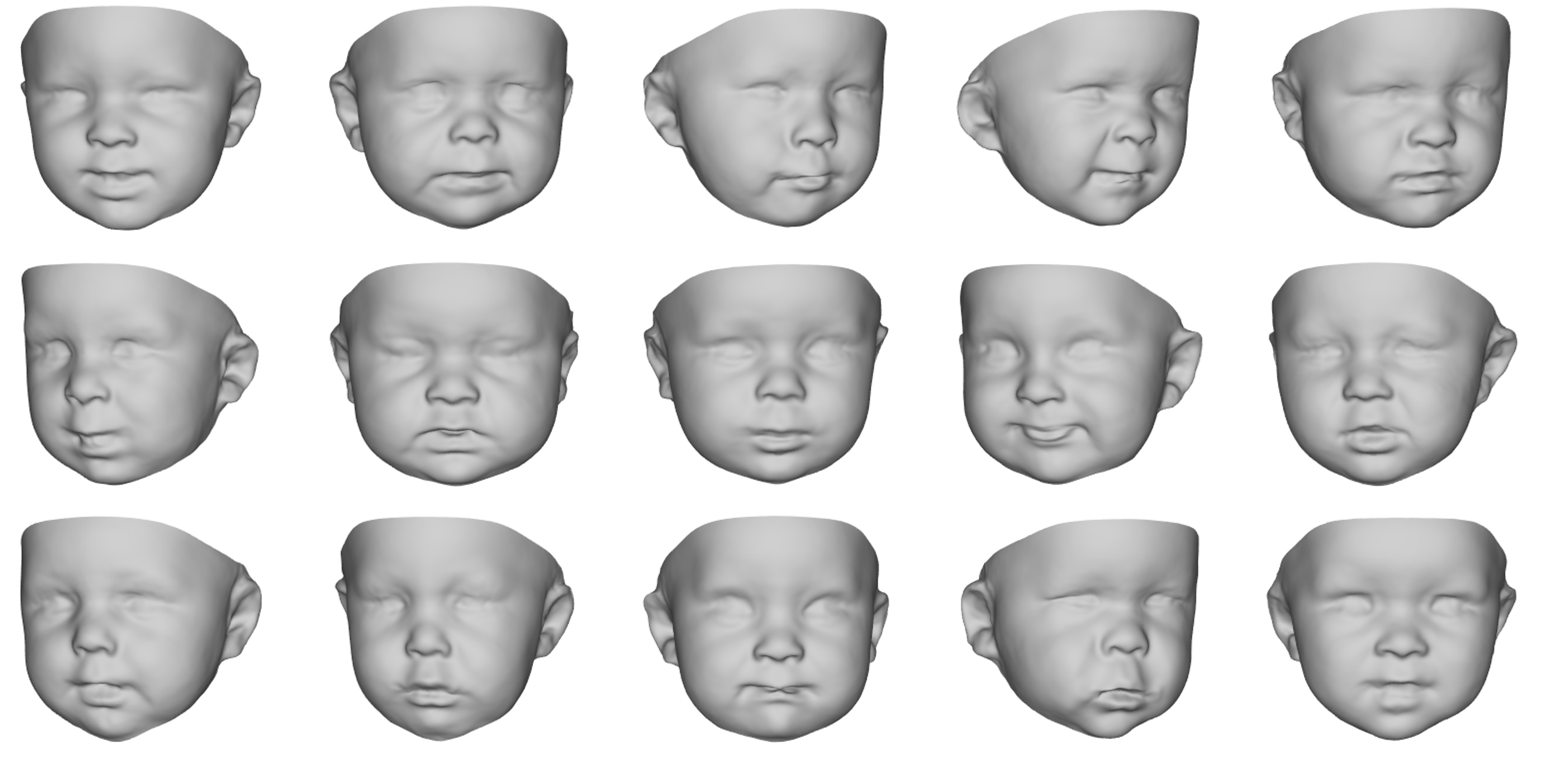}
       \caption{Examples of 3D synthetic infant faces generated with BabyFlow by randomly sampling identities and
expressions.}
      \label{fig:synthetic}
\end{figure} 

We evaluated BabyFlow’s ability to generate realistic and expressive 3D infant faces. Figure \ref{fig:synthetic} shows examples of 3D meshes generated by randomly sampling identities and expressions using the BabyFlow model. The generated samples exhibit realistic facial expressions and convincingly represent infant facial morphology. BabyFlow successfully models identity variations, including distinctive facial features such as pointy or rounded chin, and small or large forehead. It also captures a range of facial expressions, such as happiness, disgust, or surprise, which affect mainly the eyes, cheeks, and mouth regions.

\begin{figure}[ht]
  \centering
    \includegraphics[width= 0.9 \textwidth]{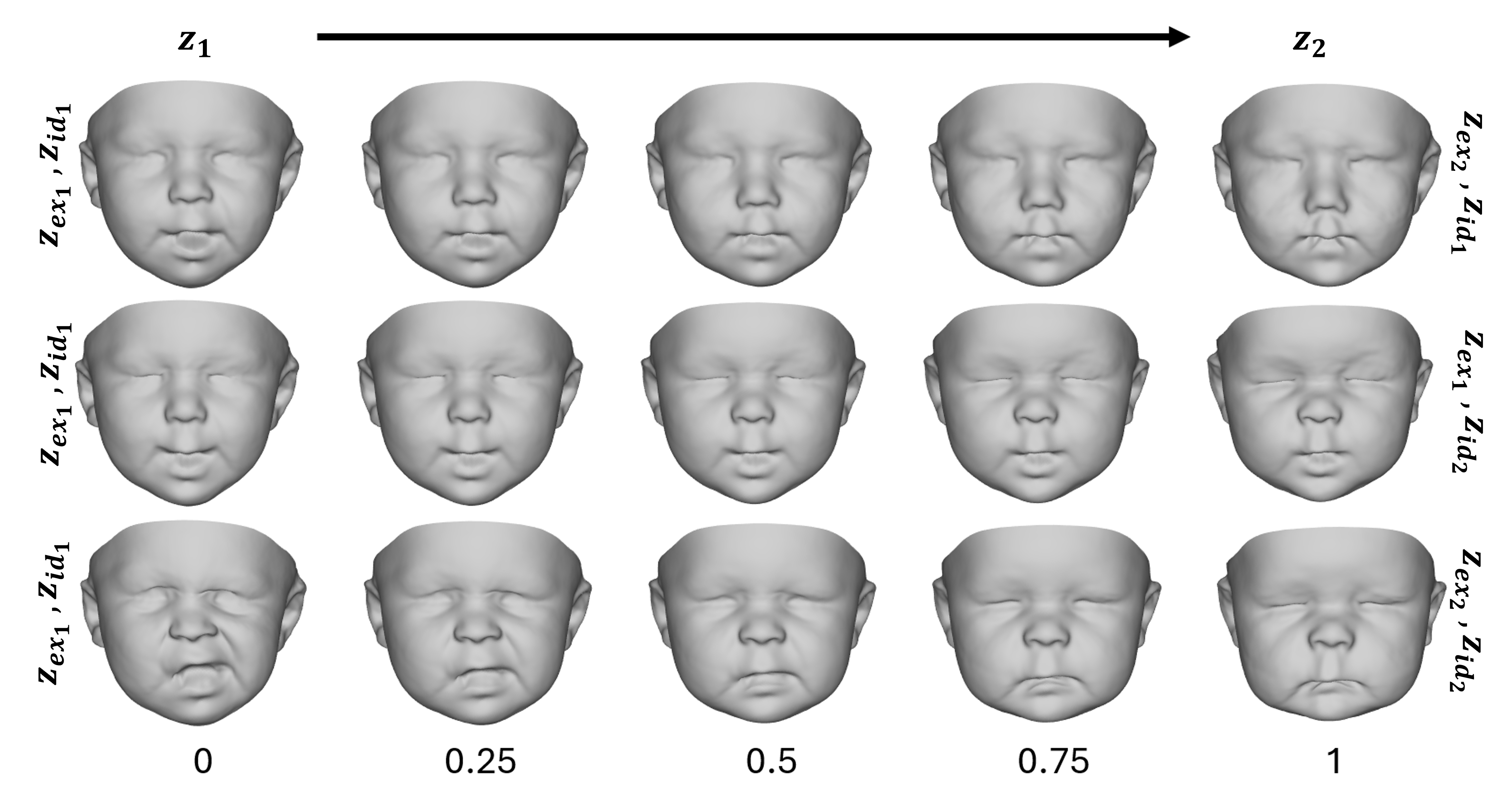}
    \caption{Latent space interpolation of the normalizing flow models. The first row varies only the expression, the second varies only the identity, and the third varies both identity and expression simultaneously. Each column represents an intermediate step in the transition, with interpolation increments of $0.25$. }     \label{fig:interpolations}
\end{figure} 

Figure \ref{fig:interpolations} demonstrates smooth latent space interpolations across different facial attributes, showcasing the model’s ability to disentangle identity and expression. In the first row, interpolation between two expressions while preserving identity results in a natural transition from a surprised to a serious facial expression, with consistent facial structure throughout. The second row illustrates interpolation between two distinct identities while maintaining a fixed expression; in this case, facial shape gradually shifts from elongated to rounder with noticeable changes such as nose size, yet the expression remains constant. In the final row, both identity and expression vary simultaneously, producing a natural transformation in both facial geometry and facial expression appearance. Across all cases, transitions are smooth, coherent, and free from visual artifacts or distortions, confirming the robustness and realism of the latent representations.


\subsection{BabyFlow can condition diffusion models for realistic 2D pictures} \label{sec:results_diff}
We evaluated the potential of BabyFlow for cross-modal face synthesis, specifically generating realistic 2D infant images from the 3D infant scans produced by the model.
\begin{figure}[!ht]
  \centering
        \includegraphics[width= 0.95\textwidth]{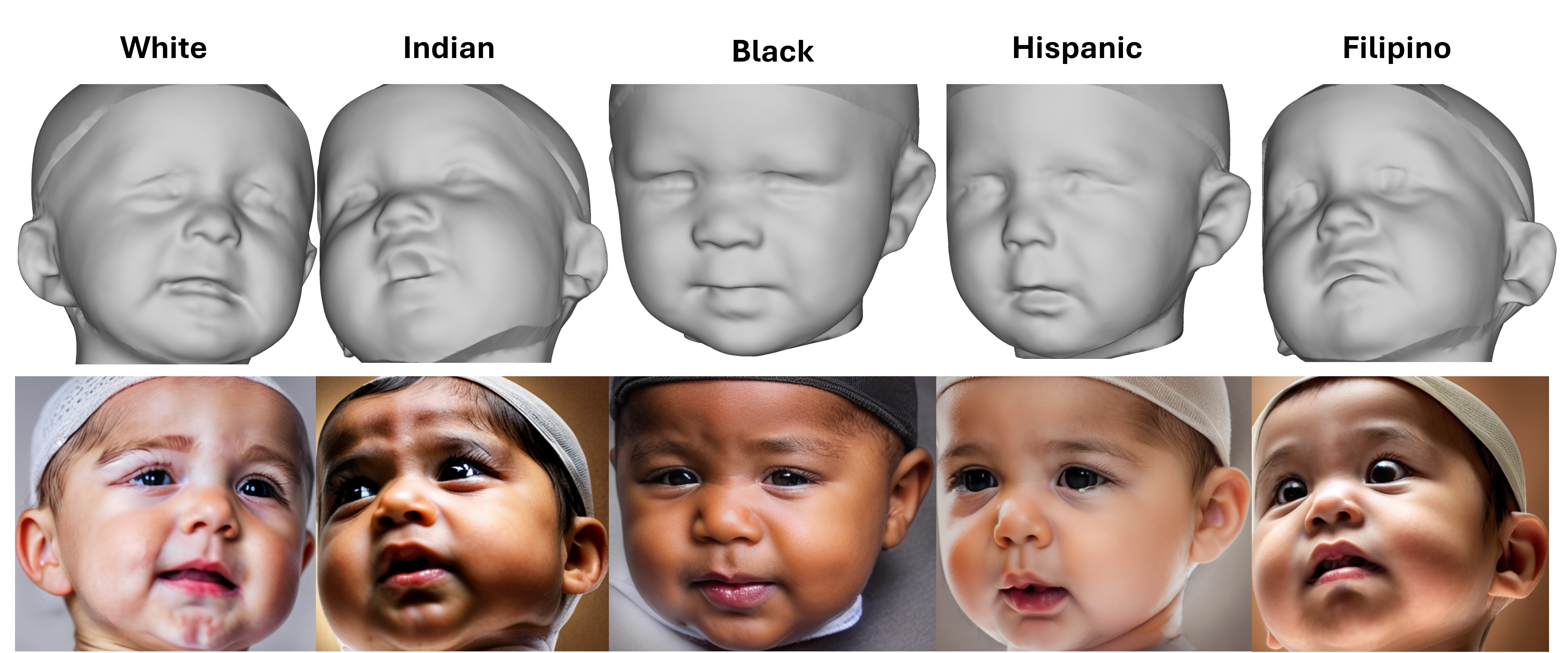}
       \caption{2D synthetic images that reflect the facial expression and morphology specified by an input 3D mesh. The top row shows randomly synthesized expressive 3D meshes generated using the BabyFlow model, while the bottom row displays the corresponding 2D synthetic images produced with SD 1.5 and guided by a multi-ControlNet to enforce consistency with the 3D infant meshes above.}
      \label{fig:synthetic_2D}
\end{figure} 

Figure \ref{fig:synthetic_2D} shows 2D synthetic images generated using the open-source diffusion model SD 1.5 \cite{Rombach_2022_CVPR} combined with a multi-ControlNet \cite{controlnet} to ensure consistency with the 3D expressive infant scans produced by the BabyFlow model. The synthesized 2D images appear realistic and expressive, and they closely reflect the facial configurations of the corresponding 3D scans. Notably, the extreme expressions present in the first and fifth columns are well preserved in the textured 2D renderings: the first column shows a clear disgust-like expression, evident in the lifted upper lip and tightened nasal area, while the fifth column conveys a surprise-like expression, with widened eyes and a slightly open mouth. These facial expressions are consistently maintained from 3D to 2D, demonstrating the model’s ability to preserve the expressive content present in the 3D shapes while adding realistic texture. The second column preserves a more relaxed configuration with gently parted lips, while the third column maintains the softly lowered eyebrows and closed lips visible in its 3D mesh. Across all columns, the 2D images preserve the overall facial geometry and the underlying facial expressions, including cheek contours, relative facial proportions, and the configuration of key expressive features, while showing expected variations in texture attributes such as skin tone, illumination, and fine detail.

\begin{figure}[ht]
  \centering
        \includegraphics[width= 0.95 \textwidth]{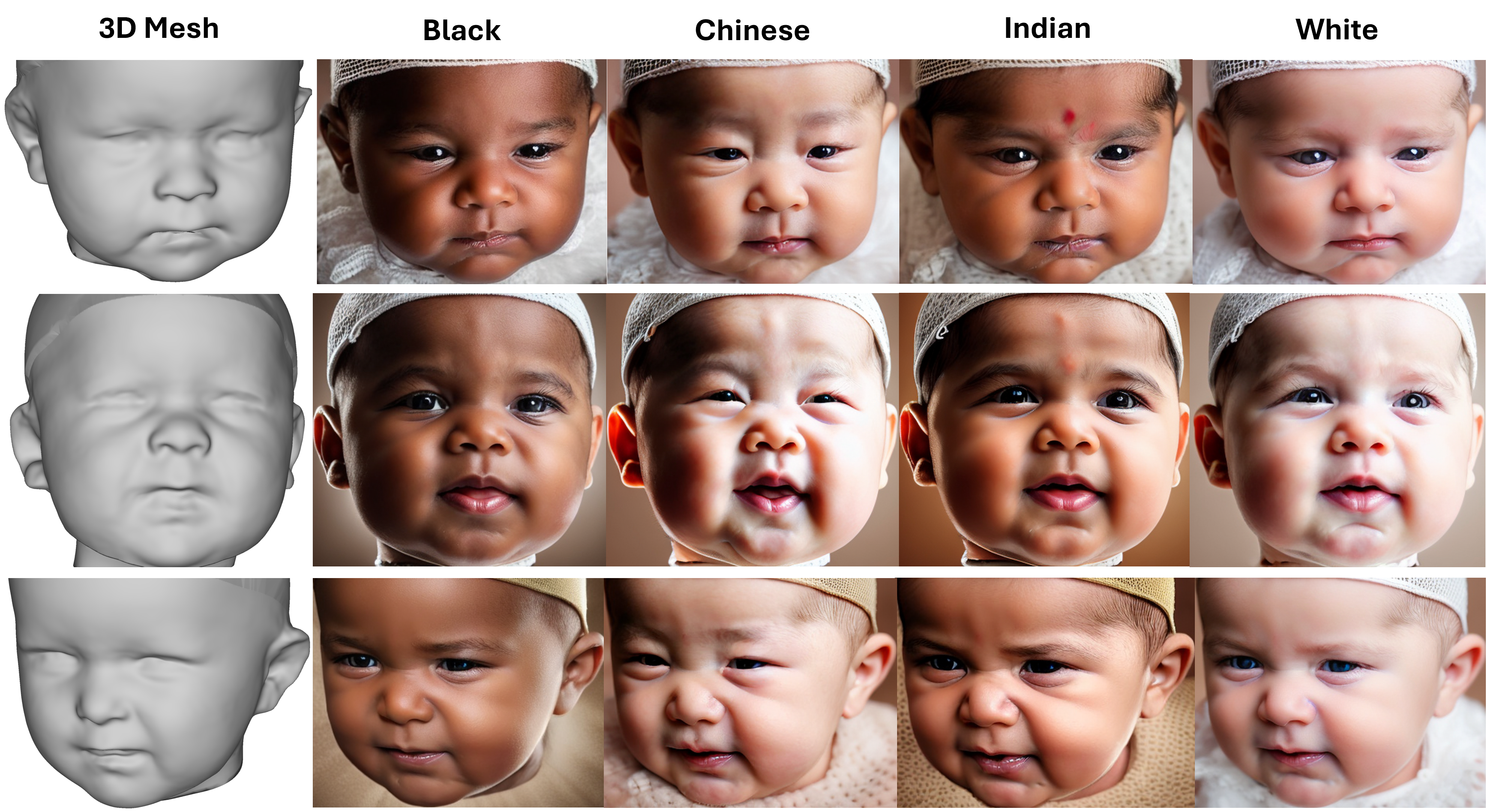}
       \caption{Multiple consistent 2D synthetic images that reflect the facial expression and morphology specified by an input 3D mesh, with variations in ethnicity controlled by the prompt. The first column shows randomly synthesized 3D meshes obtained by sampling the BabyFlow model. The remaining columns display corresponding 2D synthetic images, generated as in Figure \ref{fig:synthetic_2D}, but additionally conditioned on different ethnicities.}
      \label{fig:synthetic_2D_eth}
\end{figure} 

Figure \ref{fig:synthetic_2D_eth} shows that varying the ethnicity in the prompt produces multiple plausible 2D images that share the same 3D morphology and facial expression but differ in texture. In the top row, all generated images consistently depict a newborn face with a small, pointy chin and a serious expression, while varying in details such as skin tone, wrinkles, and illumination. A similar trend is observed in the second row, where the infant has a rounded face, scrunched nose and eyebrows, and a slightly open mouth. These morphological features are preserved across images, whereas attributes such as skin color, lip shape and thickness, and the exact degree of mouth opening vary. In the bottom row, the face shows semi-closed eyes and a scrunched nose, which remains consistent across ethnicities. For the Indian ethnicity, however, the model often introduces a small artifact on the forehead—likely reflecting a training-data bias that adds a colored dot in that region. Overall, the generated images appear plausible to the human eye. Minor variations in lip thickness and shape, or mouth opening, are present, likely due to detail loss or ambiguities introduced when projecting 3D information into 2D.

\subsection{BabyFlows learns well-behaved latent representations} \label{sec:results_latent}
A key advantage of incorporating NFs in BabyFlow lies in their ability to transform complex, high-dimensional data distributions into tractable multivariate Gaussian spaces. To assess this property, we conducted an ablation study analyzing and comparing the model's latent representations before and after applying the NFs, evaluating both the statistical behavior of the latent distributions and the quality of 3D faces generated through Gaussian sampling.

\begin{figure}[ht]
  \centering
    \begin{subfigure}[b]{0.49 \textwidth} 
    \centering    \includegraphics[width=\textwidth]{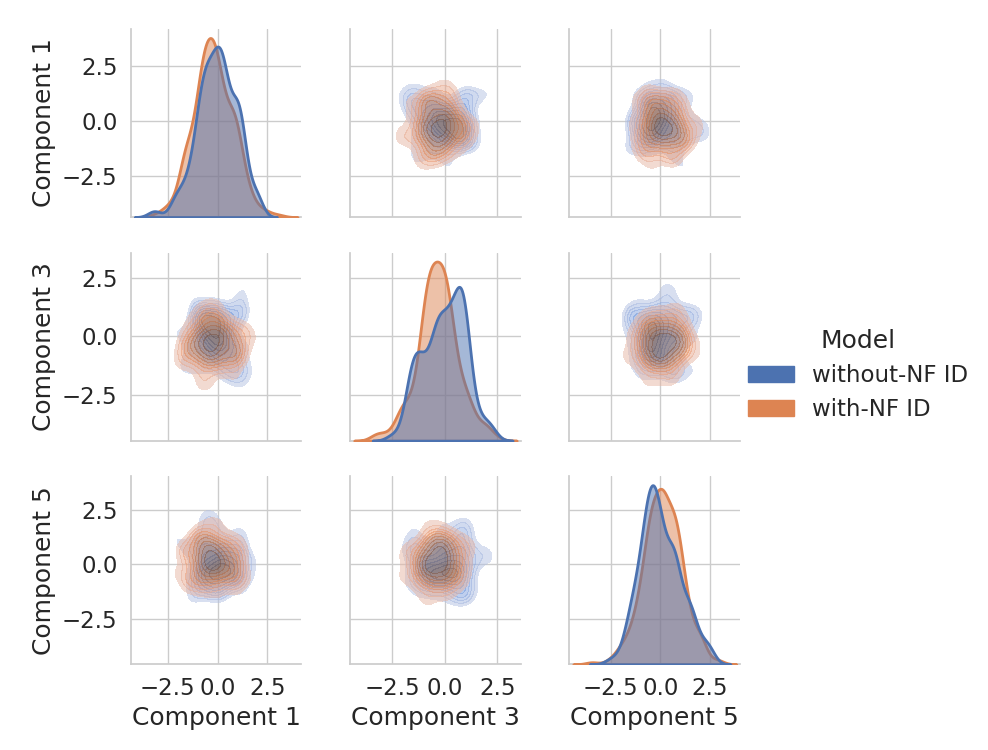}
    \caption{Identity latent space }
    \label{fig:KDE_Id_space}
    \end{subfigure}
    \begin{subfigure}[b]{0.49 \textwidth} \includegraphics[width=\textwidth]{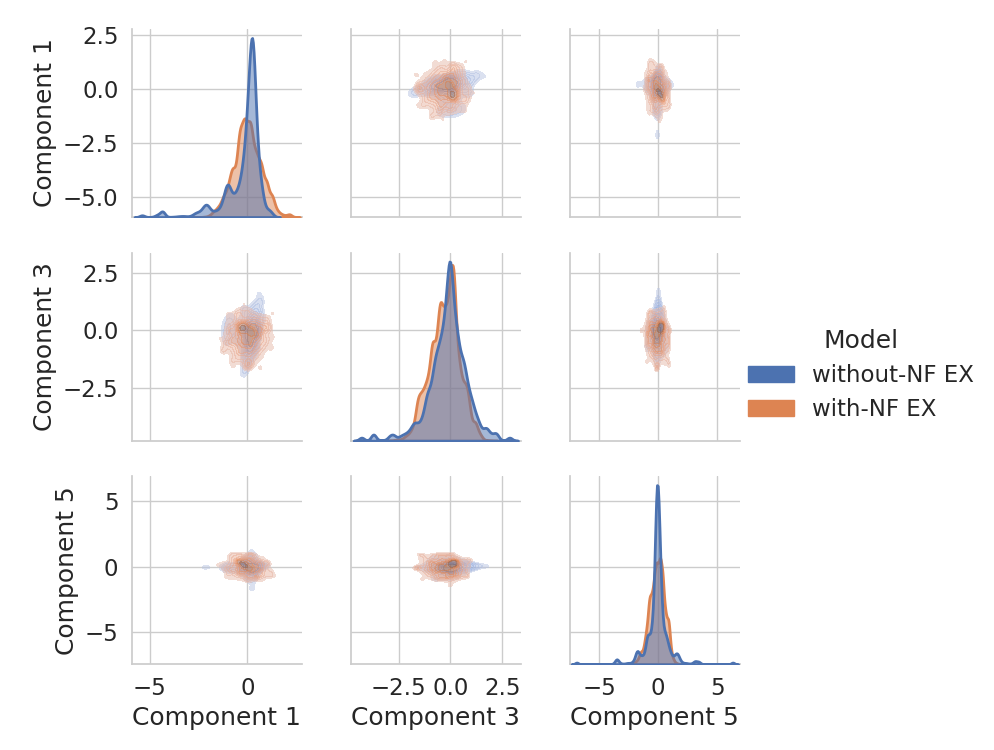}
    \caption{Expression latent space}
    \label{fig:KDE_ex_space}
    \end{subfigure}
    \caption{Paired kernel density plots of the latent space components for (a) identity and (b) expression, comparing the model before (without-NF, blue) and after (with-NF, orange) applying NFs.}
\end{figure}

Figure \ref{fig:KDE_Id_space} shows the kernel density plots of the first components of the identity latent space for both cases: before transformation with the NF (without-NF components, $\mathbf{w}_\text{id}$, in blue) and after transformation with the NF (with-NF components, $\mathbf{z}_\text{id}$, in orange). The without-NF components clearly deviate from a Gaussian distribution, whereas the with-NF components exhibit a more regular, approximately Gaussian behavior. A similar trend is observed in Figure \ref{fig:KDE_ex_space}, which illustrates the distributions of the expression latent space components: the with-NF representations ($\mathbf{z}_\text{ex}$, in orange) are smoother and closer to Gaussian than the without-NF ones ($\mathbf{w}_\text{ex}$, in blue).

\begin{figure}[ht]
  \centering
        \includegraphics[width= 0.8 \textwidth]{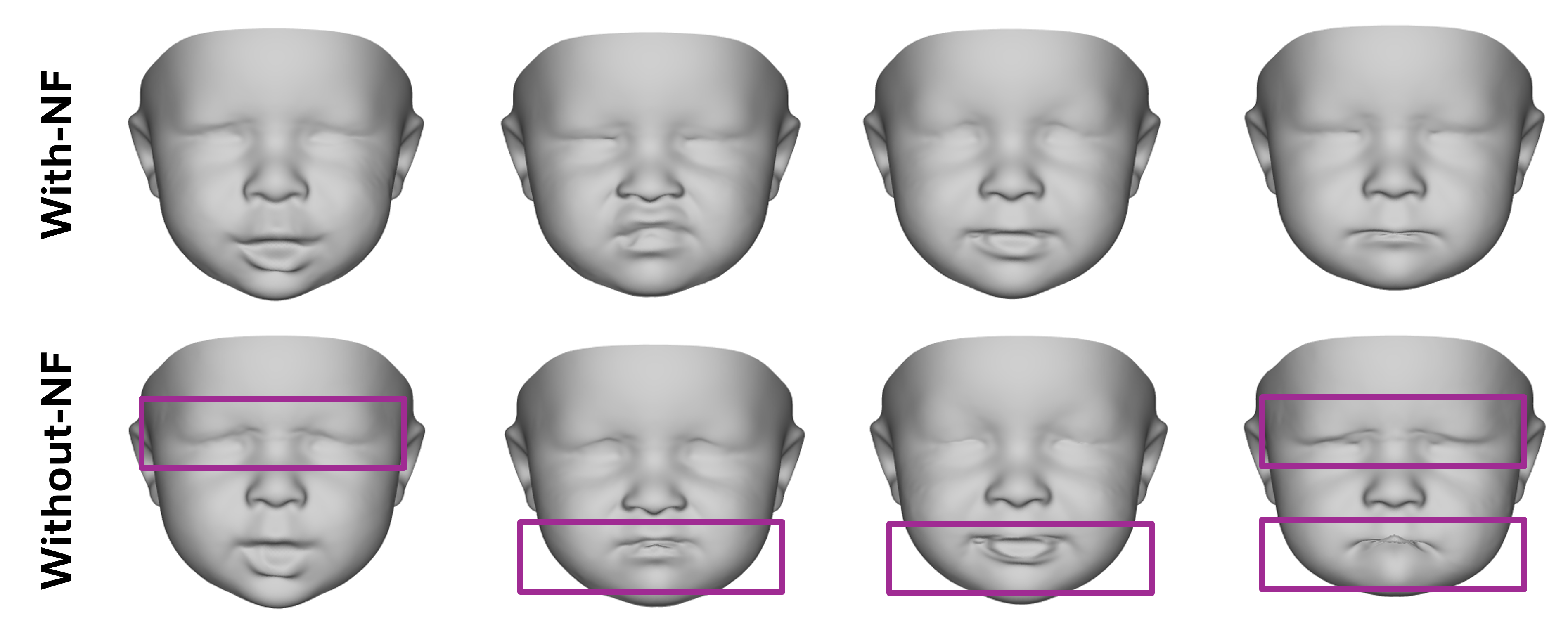}
       \caption{Synthetic expressions generated using the with-NF model (top row, $\mathbf{z}_\text{ex}$) and the without-NF model (bottom row, $\mathbf{w}_\text{ex}$). Each column corresponds to a randomly sampled expression, with identity fixed to the mean shape.
       }
      \label{fig:synthetic_ex}
\end{figure}

We further evaluated the quality of the learned latent spaces through random sampling, as described in Section \ref{sec:sampling}. Figure \ref{fig:synthetic_ex} shows 3D facial expression shapes generated by sampling from the expression latent space while keeping the identity fixed to the mean shape (${\boldsymbol\mu}_\text{id}$). The samples produced by the without-NF model (bottom row) often appear exaggerated and suffer from mesh distortions, especially around the mouth region. In contrast, the with-NF model (top row) generates more realistic and smooth expressions, with fewer artifacts. 

Analogously, Figure \ref{fig:synthetic_id} shows the result of sampling from the identity latent space, while keeping the expression fixed to the neutral configuration. Although identity variations  are more subtle, the without-NF model again produces some unrealistic samples, including poorly defined lips (bottom left), which could negatively affect tasks such as such as facial expression editing. In contrast, the with-NF model produces more coherent and anatomically plausible identities.

These results demonstrate that BabyFlow learns well-structured latent spaces, thanks to the use of NFs, which lead to more reliable sampling and generation, especially for facial expressions, and improve over the more irregular and artifact-prone behavior observed in the model without the NFs.

\begin{figure}[ht]
  \centering
        \includegraphics[width= 0.8 \textwidth]{latent_id.png}
       \caption{Synthetic identities generated using the with-NF model (top row, $\mathbf{z}_\text{id}$) and the without-NF model (bottom row, $\mathbf{w}_\text{id}$). 
       Each columns corresponds to a randomly sampled identity, with common expression fixed. The purple box highlights areas where distortions are visible.
       }
      \label{fig:synthetic_id}
\end{figure}

\section{Discussion}
In this work, we introduce BabyFlow, a flow-based generative model for accurately representing facial shape and expression in infants. BabyFlow addresses long-standing challenges in 3D infant facial analysis, including limited data availability, spontaneous expressions, and the poor transferability of adult-trained models. By disentangling identity and expression and employing NFs for both components, the model overcomes the linearity and Gaussian assumptions of traditional 3DMMs, enabling a more accurate and flexible representation of non-linear facial variations. BabyFlow supports realistic infant face generation, expression editing, and neutralization, offering strong potential for applications such as 3D face reconstruction, data synthesis for training deep learning models, and expression-invariant analysis in clinical settings.

Results from the 3D-3D fitting experiments demonstrate that BabyFlow achieves superior 3D reconstruction accuracy, particularly in expression-prone regions such as the eyes, mouth, nose, and chin (Section \ref{sec:results_3Drec}). By explicitly modeling expression variability and decoupling identity and expression parameters, BabyFlow significantly reduces reconstruction error and improves generalization to unseen data, outperforming prior work such as BabyFM \cite{Morales2025}. In addition to accurate reconstruction, BabyFlow enables controlled face generation by sampling from multivariate Gaussian distributions in the identity and expression latent spaces (Section \ref{sec:results_synth}). This allows independent manipulation of expression and identity, preserving one while modifying the other, and supports smooth interpolation in the latent space, opening the door to applications such as infant facial video reenactment. 
Overall, our results demonstrate that BabyFlow can successfully perform cross-age (adult-to-infant) expression transfer, highlighting its potential for accurate and realistic facial synthesis across age groups. We also assessed its cross-modal capabilities (geometry-to-image) by combining BabyFlow with 2D diffusion models to generate paired 3D–2D data (Section \ref{sec:results_diff}), which could facilitate training 3D reconstruction networks from 2D images. Finally, our ablation analysis reveals the key role of NFs in learning structured, stable latent spaces (Section \ref{sec:results_latent}). Without NFs, the latent representations deviate from a multivariate Gaussian and result in unrealistic faces and mesh distortions. In contrast, BabyFlow generates plausible and consistent facial geometries and expressions.

Despite its promising results, BabyFlow has some limitations. The training dataset is relatively small and may not capture the full diversity of infant facial morphology across populations, ages, and medical conditions. Moreover, while the framework shows strong technical performance, its clinical utility still requires validation in real diagnostic workflows. 
These open challenges naturally define the next steps for this line of research, which include expanding the dataset to encompass broader morphological diversity and assessing performance in real-world clinical scenarios. A particularly promising avenue is the application of its expression-neutralization capabilities to previously unusable scans, thereby broadening data inclusion without the need for neutral-expression recordings.

Even with these open challenges, BabyFlow represents a step toward infant-specific, expression-aware tools for early craniofacial assessment, with the ability to advance 3D infant facial analysis through realistic face generation, controlled editing, and expression neutralization. With continued development and validation, it could become a widely adopted resource for both research and clinical practice, enabling more inclusive and accurate analysis of infant craniofacial data.



\section{Methods} \label{sec:methods}
Figure \ref{fig:BabyX} shows the pipeline for constructing the BabyFlow model, a generative disentangled representation of identity and expression for infant faces. To achieve this, there are three main stages: 1) data pre-processing (Section \ref{sec:meth_preprocessing}), 2) disentangled representation learning of identity and expression (Section \ref{sec:meth_disentangling}), and 3) deep invertible transformation to learn well-behaved latent distributions over identity and expression spaces (Section \ref{sec:meth_NF}). In the following subsections, we first present the dataset used in this study (Section \ref{sec:meth_data}), then describe each of these three stages, and finally outline the experimental setup and implementation details (Section \ref{sec:experiments}). 

\subsection{Data} \label{sec:meth_data}
The data used to construct the BabyFlow model consist of 3D head photography scans of pediatric patients collected directly at Children’s National Hospital, Washington, DC, USA, using the 3dMDhead System (3dMD, Atlanta, GA). The study was reviewed and approved by the Institutional Review Board (IRB) of Children’s National Hospital. All methods were performed in accordance with the relevant guidelines and regulations, including the principles of the Declaration of Helsinki (1964) and its later amendments (2013). Written informed consent was obtained from the parents or legal guardians of all participating minors for the use of their data for research purposes.

It consists of 3D scans from 115 infants with non-affected facial geometries, aged between 1 and 34 months (mean age of $11.25$ months and standard deviation of $8.98$). The data is roughly gender-balanced (64$\%$ male) and includes diverse racial and ethnic backgrounds: $48\%$ Caucasian, $21\%$ Hispanic, $18\%$ African American, $8\%$ Asian, and $5\%$ undetermined. A spectral correspondence framework was used to establish the correspondences between the infant scans, as detailed in Morales et al. \cite{Morales2025}. These scans are part of the BabyFM \cite{Morales2025}, the first 3DMM of infant faces. The BabyFM includes 23 anatomical landmarks that serve as consistent and biologically meaningful reference points across subjects (see Supplementary Figure 1).

However, despite the anatomical detail and demographic diversity, the dataset lacks systematic expression variability, with just one (often non-neutral) scan per subject. To overcome this limitation and enrich the expressiveness of our model, we transfer facial expressions from FaceWarehouse (FW) \cite{Zhou2014}, a widely used 3D facial expression database of adults. This cross-age expression transfer (explained in Section \ref{sec:meth_disentangling}) is motivated by established similarities in facial expression patterns between adults and infants \cite{Ekman2002, Ekman1979, Oster1992, Oster2006}. Facial expressions are described in terms of action units (AUs) \cite{Ekman2002}, which represent fundamental facial muscle movements and provide a standardized framework for modeling facial dynamics. The FW dataset consists of 150 individuals and includes 47 blendshapes representing 46 AUs, as defined in \cite{Ekman2002}, plus a neutral expression.

To protect patient privacy, no patient facial scans are shown in the manuscript. All figures and visualizations presented are synthetic 3D facial scans or 2D images generated by the trained model and do not represent identifiable individuals.

\subsection{Pre-processing} \label{sec:meth_preprocessing}
To obtain a disentangled representation of identity and expression in infant faces, we perform expression transfer from adult to infant scans to generate a full set of infant expressions (Section \ref{sec:meth_disentangling}). Before doing so, we need to ensure compatibility between the two datasets, which we address by the following pre-processing steps: 1) mapping the mesh topology between the adult and infant scans, and 2) detecting facial expressions in the infant scans. 

\subsubsection*{Mapping between adult and infant scans}
We define a spatial mapping $\cal{\mathbf{\mathscr{M}}}$ that reparameterizes the adult facial meshes from FW using the BabyFM triangulation (which is the shared mesh topology used to represent our infant scans). This mapping must bridge significant differences in mesh resolution (11K vertices in FW vs. 36K in BabyFM) and anatomical structure (adult vs. infant morphology).

To address these challenges, we first fill the eyes and mouth holes in the FW meshes using a paraboloid-based approximation to obtain closed surfaces and stabilize subsequent alignment. Then, we perform a multi-stage fitting procedure adapted from Dai et al. \cite{Dai2020} between the average infant ($\boldsymbol{\mu}_\text{BabyFM}$) and adult ($\boldsymbol{\mu}_\text{FW}$) meshes. The fitting consists of template adaptation and iterative coherent point drift morphing, each followed by a Laplace-Beltrami regularized projection to ensure geometric consistency. A more detailed description of the fitting process is available in our previous work \cite{Alomar2022}.

The result of the previous fitting is the FW mean represented in the BabyFM triangulation, denoted as $\boldsymbol{\mu}_{\text{FW}\rightarrow\text{BabyFM}}$. To more accurately define correspondences between the two mesh domains, we represent them using barycentric coordinates rather than standard point-to-point mappings. This allows for a more flexible representation across meshes with different vertex counts, such as in our case. Specifically, once the fitted mesh
$\boldsymbol{\mu}_{\text{FW}\rightarrow\text{BabyFM}}$ is obtained, we can express the coordinates $\mathbf{p}_i$ of each vertex $i$ in the BabyFM triangulation as a barycentric combination of the FW mesh vertices:
\vspace{-3mm}
\begin{align} \label{baricentric1}
 \mathbf{p}_i = 
 \begin{bmatrix} \mathbf{v}_1 & \cdots& \mathbf{v}_M \end{bmatrix} \begin{bmatrix}
     \alpha^i_1\\
     \vdots\\
     \alpha^i_M\\
 \end{bmatrix} &= \alpha^i_1 \mathbf{v}_1 +\cdots +  \alpha^i_M \mathbf{v}_M
\\
&\cong \alpha^i_q  \mathbf{v}_q + \alpha^i_r  \mathbf{v}_r + \alpha^i_l \mathbf{v}_l \nonumber
\end{align} 
where $\mathbf{v}_j \in \mathbb{R}^{3}$ are the vertex coordinates of the FW mesh, and $\alpha^i_j$ are the barycentric weights such that $\sum^M_j \alpha^i_j=1$, with $\alpha^i_j \in [0,1]$. In practice, only three non-zero weights are retained, corresponding to the nearest FW triangular face to vertex $\mathbf{p}_i$. These three vertices $\mathbf{v}_q$, $\mathbf{v}_r$, and $\mathbf{v}_l$ define the triangle over which $\mathbf{p}_i$ is interpolated, with associated weights $\alpha^i_q$, $\alpha^i_r$, and $\alpha^i_l$.

Using this formulation, we define a linear mapping $\cal{\mathbf{\mathscr{M}}}$ from a FW mesh $\mathbf{Y}_{FW} \in \mathbb{R}^{3 \times M}$ to its BabyFM-triangulated counterpart $\mathbf{Y}_{\text{FW}\rightarrow\text{BabyFM}} \in \mathbb{R}^{3 \times N}$ as:
\begin{align} \label{baricentric}
&\mathbf{Y}_{\text{FW}\rightarrow\text{BabyFM}} = \mathbf{Y}_\text{FW} \cdot\cal{\mathbf{\mathscr{M}}} 
 \\
 &= \begin{bmatrix}
     \mathbf{v_1} & \mathbf{v_2} &  \cdots & \mathbf{v_M}
 \end{bmatrix} \begin{bmatrix}
     \alpha^1_1 & \alpha^2_1 & \cdots & \alpha^N_1\\
     \alpha^1_2 & \alpha^2_2 & \cdots & \alpha^N_2\\
     \vdots & \vdots & \ddots & \vdots\\
     \alpha^1_M & \alpha^2_M & \cdots & \alpha^N_M\\
 \end{bmatrix}. \nonumber 
\end{align}
Here, each column of $\cal{\mathbf{\mathscr{M}}}$ contains the barycentric coefficients that express one vertex of the BabyFM-triangulated mesh as an affine combination of three vertices from the FW mesh triangle in which it lies. This representation enables a smooth and resolution-agnostic mapping, allowing us to transfer expressions between meshes with different vertex counts while preserving spatial accuracy and continuity across domains.

\subsubsection*{Facial expression recognition}
Our 3D infant scans contain arbitrary expressions, typically non-neutral, which introduces significant geometric deformations. If not properly accounted for, these deformations can interfere with the expression transfer process by compounding with the newly applied expression-induced changes, leading to unrealistic or distorted facial geometries. To mitigate this, we first detect the existing expressions  so that they can be factored in during expression transfer. 

To perform expression recognition, we follow the approach of Derkach et al. \cite{Derkach2018} based on local shape spectral analysis. Since facial expressions are primarily located in the frontal part of the face, we compute the spectral representation of local patches centered on the 19 central anatomical landmarks of the BabyFM (see Figure \ref{fig:lmks} of supplementary material for illustration of all the 23 landmarks). For each landmark $k$, we define a mesh patch $\mathbf{\mathscr{P}}=(\mathscr{V},\mathscr{E})$, where $\mathscr{V}$ is the set of $n$ vertices and $\mathscr{E}$ the edge connections of the 1-ring neighborhood. We compute the graph Laplacian $ \mathbf{L}_k$ as an $n \times n$ matrix defined by a discrete Schr\"odinger operator as:
\begin{equation} \label{Graph_Laplacian}
[\mathbf{L}_k]_{ij}= \left\{ \begin{array}{lcc}
             -1 &  \text{if}  & (i,j) \in \mathscr{E}\\
             \xi_i & \text{if}  & i=j\\
             0 & \text{otherwise} ,
             \end{array}
   \right.
\end{equation} 
where $\xi_i$ is the valence or degree of vertex $i$.

We perform eigen-decomposition of $\mathbf{L}_k$ to obtain a local spectral representation of the spatial information around every landmark $k$. Then, we define the $\tau$-dimensional embedding $\Phi^{k}_\tau  = [\boldsymbol{\phi}_1 \boldsymbol{\phi}_2 \ldots \boldsymbol{\phi}_\tau]$ to represent the global patch structure of landmark $k$. This embedding contains the eigenvectors $ \boldsymbol{\phi}_1, \ldots , \boldsymbol{\phi}_\tau$ associated with the $\tau$ smallest eigenvalues, which represent the low-frequency information of the local geometry. By removing high frequencies, this representation becomes more robust to local noise. The mesh coordinates of each landmark patch can then be projected into the spectral domain to obtain the spectral feature vector $\mathbf{\Omega_{v_k}} \in \mathbb{R}^{\tau}$ as:
\begin{equation} \label{projection}
   \mathbf{\Omega_{v_k}} =  (\Phi^{k}_{\tau})^{T} \mathbf{v}_k
\end{equation}
where $\mathbf{v}_k \in \mathbb{R}^n$ denotes the patch coordinates of landmark $k$. 
We describe each mesh using the feature matrix $\mathbf{\Omega_v} = [\mathbf{\Omega_{v_1}} \mathbf{\Omega_{v_2}} \ldots  \mathbf{\Omega_{v_{19}}}]$, where each $\mathbf{\Omega_{v_k}}$ is a $\tau$-dimensional embedding with $\tau = 50$. This embedding dimension has been shown to perform well for facial expression recognition \cite{Derkach2018}. Then, we use these embeddings to train 46 binary support vector machine (SVM) classifiers using the LIBSVM library \cite{chang2011libsvm}. These classifiers are designed to identify the AUs present in the original (typically non-neutral) infant scans. The output is a 46-dimensional vector, where each entry is set to 1 if the corresponding AU is detected, and 0 otherwise.

\subsection{Disentangled representation learning of identity and expression in infant faces}\label{sec:meth_disentangling}
After pre-processing, the second stage of the pipeline (Figure \ref{fig:BabyX}) is to obtain a decoupled representation of identity and expression. To this end, we construct a bilinear model \cite{Vlasic2005}. However, this requires assembling a data tensor that contains a complete set of expressions for each identity. 

\subsubsection*{Cross-age expression transfer} \label{sec:cross-age_ex}
To transfer an expression $e$ to an infant scan with expression $s$, represented by $\mathbf{X}_{\text{BabyFM},s} \in  \mathbb{R}^{3 \times N}$, we adopt a methodology inspired by our previous work  \cite{Alomar2022}. Specifically, we randomly select an individual from the FW database and retrieve their facial geometry under both the source $s$ and the target $e$ expressions, denoted as $\mathbf{Y}_{\text{FW},s}$ and $\mathbf{Y}_{\text{FW},e} \in  \mathbb{R}^{3 \times M}$, respectively. The detected expression $s$ in the infant scan (Section \ref{sec:meth_preprocessing}) may be a combination of AUs, which we similarly combine to synthesize the corresponding adult expression. The deformation vector in the FW triangulation from $s$ to $e$ is obtained as:
 \begin{equation}  \label{def_vector}
    \mathbf{D}_\text{FW} = \mathbf{Y}_{\text{FW},e} - \mathbf{Y}_{\text{FW},s} \, .
 \end{equation}
Then, the computed mapping $\cal{\mathbf{\mathscr{M}}}$ (Section \ref{sec:meth_preprocessing}) allows us to represent $\mathbf{D}_\text{FW}$ within the BabyFM framework, as follows:
\begin{equation}  \label{def_vector_baby}
\mathbf{D}_{\text{FW} \rightarrow \text{BabyFM}} = \mathbf{D}_\text{FW} \cdot \cal{\mathbf{\mathscr{M}}} 
\end{equation}
where $\mathbf{D}_{\text{FW}\rightarrow\text{BabyFM}} \in  \mathbb{R}^{3 \times N}$. This transformed deformation can then be applied to the infant mesh $\mathbf{X}_{\text{BabyFM},s}$ to generate a new mesh $\mathbf{X}_{\text{BabyFM},e} \in  \mathbb{R}^{3 \times N}$ with the transferred expression $e$:
\begin{equation} \label{eq:face_transfer}
    \mathbf{X}_{\text{BabyFM},e} =  \mathbf{X}_{\text{BabyFM},s} + \delta \  \mathbf{D}_{\text{FW} \rightarrow \text{BabyFM}}
\end{equation}
where $\delta$ is a scalar that controls the intensity of expression transfer. 

To minimize the dependency on any individual, we apply an ensemble-based strategy: the expression transfer process is repeated for $\kappa = 40$ randomly selected subjects from the FW database, and the resulting meshes are averaged.



Out of the 115 meshes, 95 allowed for a successful expression transfer and were therefore kept for constructing the BabyFlow model. The remaining 20 scans were used as test set for testing the 3D-3D fitting with the resulting model. The transferred facial expressions are the 46 AUs available in the FW database, the 6 universal expressions, the compound expressions \cite{ShichuanDu2015}, and some characteristic infant facial expressions (e.g., crying with closed eyes), totalling $n_\text{ex}=76$ expressions. The detailed list of facial expressions considered to form the infant data tensor is provided in Supplementary Table 1 and 2. 

\subsubsection*{Bilinear infant model}
Once expression transfer is completed, all meshes are correctly aligned to remove translation and rotation, as these components are irrelevant to facial morphology. To augment the dataset and exploit the near-symmetric topology of the face mesh, we generate both a mirrored and a symmetrized version of each infant face shape, thereby introducing reflected and averaged-symmetry variants. The resulting dataset is organized into a 3-mode data tensor $ \mathbf{T} \in \mathbb{R}^{3N \times 3 n_\text{id} \times n_\text{ex}}$, where $N$ is the number of vertices of the BabyFM triangulation, $n_\text{id}$ is the number of infant identities, and $n_\text{ex}$ is the number of facial expressions. Thus, the tensor has dimensions $93 098 \times 285 \times 76$, encoding the shape variations across all combinations of identity and expression. 

We apply higher-order singular value decomposition (HOSVD) to decompose the tensor, obtaining the following representation:
\begin{equation}  \label{HOSVD}
    \mathbf{T} = \mathbf{S} \otimes \mathbf{U}_\text{verts} \otimes \mathbf{U}_\text{id} \otimes \mathbf{U}_\text{ex}
\end{equation}
where $\mathbf{S}$ is the core tensor, the $\mathbf{U}$ matrices are the orthogonal bases for the corresponding subspaces, and $\otimes$ denotes the outer product. This decomposition allows us to model and decouple identity and expression. Thus, a 3D face $\mathbf{x}\in \mathbb{R}^{3N}$ can be described as:
\begin{equation}  \label{eq:bilinear} 
    \mathbf{x} = \mathbf{C} \otimes  \mathbf{w}_\text{id} \otimes \mathbf{w}_\text{ex} +  \boldsymbol{\mu_x}
\end{equation}
where $\mathbf{C} = \mathbf{S} \otimes \mathbf{U}_\text{verts}$ is the core tensor projected onto the vertex subspace, $\boldsymbol{\mu_x}$ is the mean of the bilinear model, and $\mathbf{w}_\text{id}$ and $\mathbf{w}_\text{ex}$ are the identity and expression parameters, respectively.

\subsection{Deep invertible transformation to learn well-behaved latent distributions over identity and expression spaces} \label{sec:meth_NF}

The bilinear model in Equation \ref{eq:bilinear} provides a compact and interpretable latent representation of 3D facial geometry, where each face $\mathbf{x} \in \mathbb{R}^{3N}$ is encoded by an identity vector $\mathbf{w}_\text{id} \in \mathbb{R}^{n_\text{id}}$ and an expression vector $\mathbf{w}_\text{ex} \in \mathbb{R}^{n_\text{ex}}$. However, while this factorization allows the separation of identity and expression, it does not offer a principled way to model the underlying distributions of these latent variables, which are often assumed to follow a Gaussian prior, an assumption that may not hold in practice. To enable generative capabilities, exact likelihood estimation, and improved control over identity and expression, we learn structured latent spaces over $\mathbf{w}_\text{id}, \mathbf{w}_\text{ex}$ using deep invertible transformations, specifically NFs \cite{Kobyzev2021}. This approach captures the complex, non-linear distributions of identity and expression while preserving the interpretability provided by the bilinear structure.

\begin{figure*}[ht]
  \centering
        \includegraphics[width= 1 \textwidth]{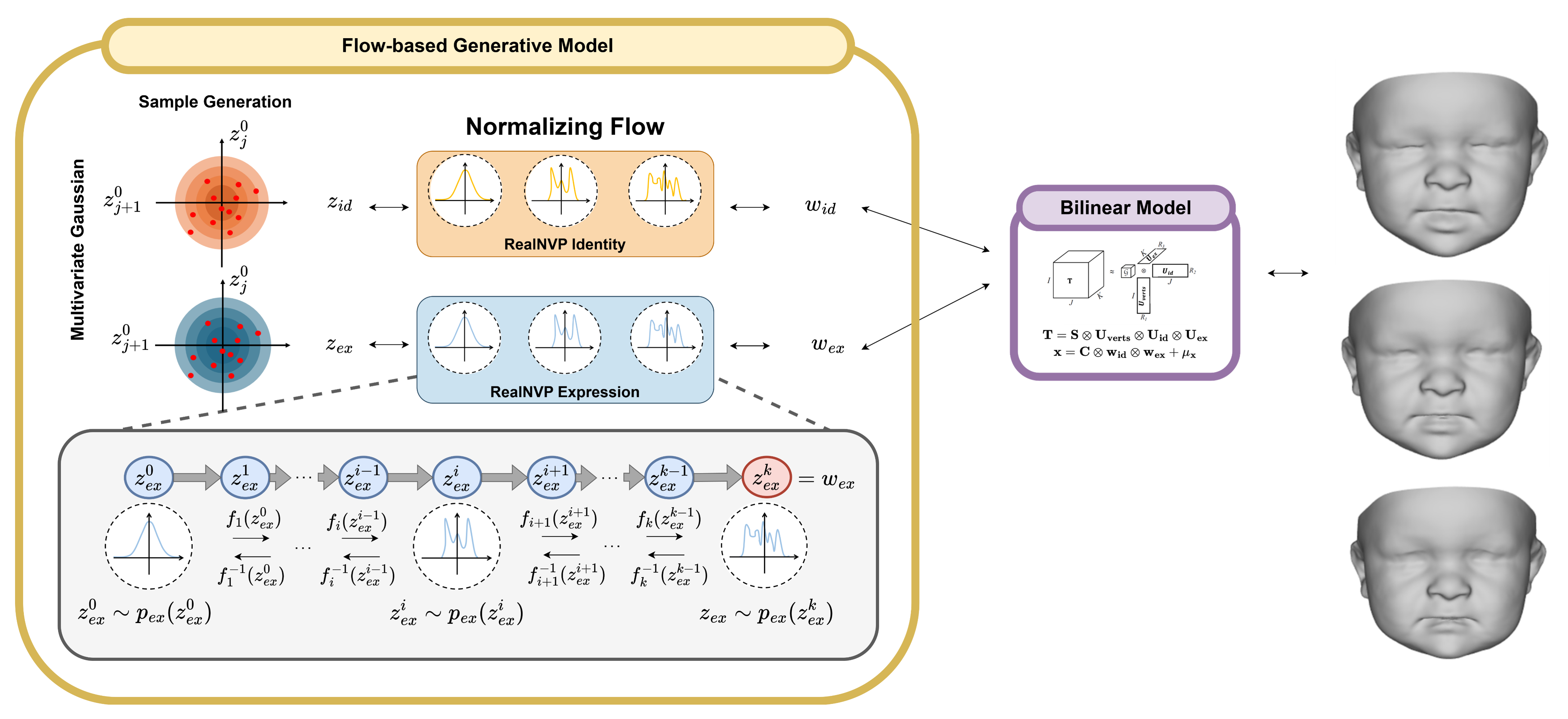}
        \vspace{-2mm}
       \caption{Flow-based generative model. The model enables sampling of new identities and expressions from multivariate Gaussian distributions by using two normalizing flow blocks that accurately capture the true, non-Gaussian density of the latent spaces. The sampled identity and expression codes are then combined via a bilinear model to generate 3D representations of expressive infant facial shapes.}
      \label{fig:BabyFlow}
\end{figure*} 

\subsubsection*{Mathematical framework}
Let $\mathbf{w} \in \mathbb{R}^{D} $ be a random variable (corresponding to either $\mathbf{w}_\text{id}$ or $\mathbf{w}_\text{ex}$), with a probability density function $p_W(\mathbf{w})$, referred to as target distribution, and let $\mathbf{z} \in \mathbb{R}^{D}$ be a random variable drawn from a well-defined base distribution $p_Z(\mathbf{z})$, typically a multivariate Gaussian. An NF is a bijective (and therefore invertible) differentiable function $f:\mathbb{R}^{D} \rightarrow{} \mathbb{R}^{D}$ such that $\mathbf{z}=f(\mathbf{w})$. The function $f$ is parameterized by a set of variables denoted as $\theta$, written as $f(\mathbf{w}; \theta)$, but for simplicity, we often omit the explicit dependence on $\theta$. By the change of variable theorem, the density $p_W(\mathbf{w})$ is given by:
\begin{equation}
p_W(\mathbf{w}) = p_Z(f(\mathbf{w})) \left| \det J_{f}(\mathbf{w}) \right|
\end{equation}
where $J_f(\mathbf{w})=\frac{\partial f(\mathbf{w})}{\partial \mathbf{w}}$ is the Jacobian of the transformation $f$ evaluated at $\mathbf{w}$.

The learnable parameters $\theta$ of the transformation $f$ are optimized through maximum likelihood estimation, which corresponds to minimizing the negative log-likelihood (NLL): 
\begin{align} 
\theta^* &= \arg\min_{\theta} \; \text{NLL}(\theta) \\
\text{where} \quad \text{NLL}(\theta) &= -\log  p_W(\mathbf{w}) \label{eq:NLL}  \\ 
&= -\log p_Z \left( f(\mathbf{w}) \right) - \log \left| \det J_{f}(\mathbf{w}) \right|. \nonumber
\end{align}

To model complex target distributions, a NF is typically constructed by composing multiple simpler invertible functions $f_k$, $k=1, \ldots, K$, since invertible functions are closed under composition: 
\begin{equation}
f = f_K \circ f_{K-1} \circ \cdots \circ f_2 \circ f_1 \, .
\end{equation}
This composition allows the flow to progressively transform $\mathbf{w}$ into $\mathbf{z}$, while maintaining tractable computation of the Jacobian determinant required for the NLL.

In contrast to other generative models, such as autoregressive or diffusion models, NFs provide faster training and sample generation, while also enabling exact likelihood estimation, making them well-suited to our objective. Thanks to the invertibility of the transformation $f$, sample generation is straightforward: starting from a randomly sampled point $\mathbf{z}_0 \sim p_Z(\mathbf{z})$  (i.e., drawn from a multivariate Gaussian), a new sample is  obtained by simply applying the inverse mapping: $\mathbf{w}_0 = f^{-1}(\mathbf{z_0})$. 

We implemented two NFs using the Real Non-Volume Preserving (Real-NVP) architecture \cite{Dinh2017}: one flow to model expression and another to model identity (see Figure \ref{fig:BabyFlow}). Each flow consists of a stack of 6 affine coupling layers (described below). Rather than working directly with the full-dimensional vectors, we reduced the dimensionality of each flow by retaining the principal components that capture 95\% of the cumulative variance, which preserves most of the relevant information while removing redundancy and high-frequency noise. This resulted in selecting $D=136$ for the identity NF and $D=22$ for the expression NF.

\subsubsection*{Architecture details: affine coupling layers}\label{sec:affine_coupling}
We used the Real-NVP architecture, in which the flow $f$ is implemented as a sequence of affine coupling layers \cite{Dinh2017}. In each coupling layer, a subset of the input variables is left unchanged, while the remaining variables are transformed using a scale and shift operation. The parameters for this transformation are learned by by a neural network that takes the unchanged variables as input. Thus, for the first forward transform, the input variable $\mathbf{w} \in \mathbb{R}^D$ is split into $\mathbf{z}^{(0)}_{1:d} = \mathbf{w}_{1:d}$ and $\mathbf{z}^{(0)}_{d+1:D}=\mathbf{w}_{d+1:D}$, with $d<D$, and the output of the coupling layer can be written as:
\begin{equation}\label{eq:affine_coupling}
\begin{aligned}
    \mathbf{z}^{(1)}_{1:d} &= \mathbf{z}^{(0)}_{1:d}, \\
    \mathbf{z}^{(1)}_{d+1:D} &= \exp(s(\mathbf{z}^{(0)}_{1:d})) \odot \mathbf{z}^{(0)}_{d+1:D} + t(\mathbf{z}^{(0)}_{1:d}),
\end{aligned}
\end{equation}
where $s$ and $t$ denote the learnable scale and translation functions, respectively, and $\odot$ represents element-wise product. In the subsequent coupling layer, the roles of the two partitions are swapped: the transformed variables $\mathbf{z}^{(1)}_{d+1:D}$ are left unchanged ($\mathbf{z}^{(2)}_{d+1:D}=\mathbf{z}^{(1)}_{d+1:D}$) and used to estimate the affine transformation factors to be applied to $ \mathbf{z}^{(1)}_{1:d}$, resulting in the updated partition $\mathbf{z}^{(2)}_{1:d}$. This alternating partitioning ensures that all dimensions are progressively updated across layers. Repeating this process for all coupling layers (six in our case) yields the final latent representation $\mathbf{z}=(\mathbf{z}^{(6)}_{1:d},\mathbf{z}^{(6)}_{d+1:D})=f(\mathbf{w})$.

Because each coupling layer is invertible by design, the inverse pass to map a latent variable $\mathbf{z}$ back to $\mathbf{w}$ (i.e.,$\mathbf{w}=f^{-1}(\mathbf{z})$) is obtained by simply reversing the order of layers and applying the inverse of each affine transformation. For instance, for inverting Equation \ref{eq:affine_coupling}:
\begin{equation}
\begin{aligned} \label{eq:inv_affine_coupling}
    \mathbf{z}^{(0)}_{1:d} &= \mathbf{z}^{(1)}_{1:d}, \\
    \mathbf{z}^{(0)}_{d+1:D} &= \left( \mathbf{z}^{(1)}_{d+1:D} - t(\mathbf{z}^{(1)}_{1:d}) \right) \odot \exp(-s(\mathbf{z}^{(1)}_{1:d})).
\end{aligned}  
\end{equation}
The use of affine coupling layers simplifies the Jacobian of the transformation to a lower triangular matrix. For example, for the affine coupling layer defined in Equation \ref{eq:affine_coupling}, the Jacobian is
\begin{equation}
J = 
\frac{\partial \mathbf{z}^{(1)}}{\partial \mathbf{z}^{(0)}} =
\begin{bmatrix}
I_d & 0 \\
\frac{\partial \mathbf{z}^{(1)}_{d+1:D}}{\partial \mathbf{z}^{(0)}_{1:d}} & \operatorname{diag}\!\left(\exp(s(\mathbf{z}^{(0)}_{1:d}))\right)
\end{bmatrix},
\end{equation}
where $I_d$ is the $d \times d$ identity matrix. Since $J$ is lower triangular, its determinant is simply the product of the diagonal entries, and the logarithm of the Jacobian determinant for the coupling layer reduces to the sum of the scaling values:
\begin{equation}
\log \left| \det J \right| = \sum_{j=d+1}^D s_j(\mathbf{z}^{(0)}_{1:d}) \, .
\end{equation}
This makes the computation of $\log \left|\det J_f (\mathbf{w})\right|$ in Equation \ref{eq:NLL} straightforward, as it amounts to summing the log-determinants of the six coupling layers. 

\subsubsection*{Dequantization: adding noise to smooth input and improve likelihood estimations}
NFs model continuous distributions and struggle with discrete data, as discrete inputs lead to degenerate densities and poor likelihood estimates \cite{Kobyzev2021}. To overcome this, small amounts of noise are typically added (dequantization), converting discrete data into continuous form and enabling more accurate density modeling \cite{hoogeboom2020learningdiscretedistributionsdequantization, ho19a}. Although our data are already  continuous, adding controlled noise helps smooth the distribution, which is useful given our limited training data. We therefore add random uniform noise during training to make the model more robust. Thus, given $\mathbf{w}$ (latent representation of either expression or identity before applying the NF), the dequantized version can be formulated as $\tilde{\mathbf{w}} = \mathbf{w} + \epsilon $ where $\epsilon \sim \sqrt{\sigma_{w}} U(0,1)$, with $U(0,1)$ being the uniform distribution over the interval $[0,1]$ and $\sigma_w$ being the variance of $p_W(\mathbf{w})$. 

\subsubsection*{Training objective with Jacobian regularization}
As previously mentioned, one of the advantages of NF is the simplification of the training process, as they only require minimizing the negative log-likelihood (NLL, Equation \ref{eq:NLL}). Unlike VAEs, which rely on approximate posteriors, or GANs, which depend on discriminator networks, NFs can be trained directly by optimizing the NNL. 
However, during training, certain challenges arise: the number of function evaluations and the Frobenius norm of the Jacobian matrices tend to increase \cite{Josias2022}. To prevent excessive deformation of the learned space and to aid model convergence, we regularize the Jacobian matrices using the Frobenius norm $\|J_f\|_F$, defined as:
\begin{equation}
\|J_f\|_F = \sqrt{\sum_{i,j=1}^D  |J_f^{i,j}|^2} .
\end{equation}
where $J_f^{i,j}$ denotes the entry in the $i$-th row and $j$-th column of the Jacobian matrix $J_f$. 
Adding $\|J_f\|_F$ to the loss function penalizes large Jacobian values, effectively limiting the local deformation of the transformation. This has a similar effect to controlling the spectral norm $\|J_f\|_2$, since $\|J_f\|_2 \leq \|J_f\|_F$. Therefore, our loss to train the NFs is:
\begin{align}
\mathcal{L} &= \text{NLL} + \|J_f\|_F \\
&= -\log p_Z (f(\mathbf{w})) - \log \left| \det J_f \right| + \|J_f\|_F \nonumber \, .
\end{align} 

\subsection{Experimental details} \label{sec:experiments}
Next, we describe the experiments performed to assess BabyFlow and obtain the results from Section \ref{sec:results}. 
\subsubsection*{Asessing cross-age expression transfer from adults-to-infants}
To evaluate BabyFlow’s ability to transfer facial expressions from adults to infants, 
we randomly selected a neutral infant face from our baby dataset and a neutral adult face from the FW dataset \cite{Zhou2014}. We then generated a set of target facial expressions (e.g., happy, disgust) using the FW model and applied expression transfer described in Section~\ref{sec:cross-age_ex}. This setup allows us to assess how well the model can transfer expressions from adults to children while preserving the infant’s facial identity and reproducing the expression without introducing distortions or unrealistic changes.

\subsubsection*{Asessing 3D reconstruction errors under expressive facial variation}
To evaluate the BabyFlow's ability to represent unseen subjects, we reconstructed a set of 20 3D infant scans that were not part of the training data. Each scan, represented as a 3D facial mesh $\mathbf{x} \in \mathbb{R}^{3N}$, was first rigidly aligned to the mean of the bilinear model $\boldsymbol{\mu_x}$ using Procrustes analysis. We then performed a non-linear optimization procedure with regularization to fit the BabyFlow model to the input scan. At each optimization step, we alternated between estimating the identity parameters $\mathbf{z}_\text{id}$ and the expression parameters $\mathbf{z}_{\text{ex}}$. The fitting process was initialized by setting $\mathbf{z}_\text{ex}$ to the neutral expression and $\mathbf{z}_\text{id}$ to the mean identity vector. 

The shape parameters of the model were obtained using the following non-linear optimization:
\begin{equation} \label{eq:nlo}
    E(\mathbf{z}) = \gamma_1 E_\text{verts}(\mathbf{z}) + \gamma_2 E_\text{prior}(\mathbf{z}) \,,
\end{equation}
where $\gamma_1$ and $\gamma_2$ are the corresponding weights of each error term, satisfying $\gamma_1 + \gamma_2=1$, and $\mathbf{z}=(\mathbf{z}_\text{id}, \mathbf{z}_\text{ex})^T$ are the latent parameters being optimized. The term $E_\text{verts}$ measures the reconstruction error of the vertices of the mesh, while $E_\text{prior}$ acts as a regularizer, encouraging the solution to remain within a statistically plausible region of the latent space. This prior is modeled as a multivariate Gaussian estimated from the training data, effectively constraining the solution to lie within a learned hyperellipsoid. The explicit formulations for each term in Equation \ref{eq:nlo} are:

\begin{align}
E_\text{verts} (\mathbf{z}) &=\big\| (C \otimes \mathbf{z}_\text{id} \otimes \mathbf{z}_\text{ex}) - (\mathbf{x} - \boldsymbol{\mu_x}) \big\|^2 
 \\
E_{\text{prior}}(\mathbf{z}) &= \mathbf{z}^\top \boldsymbol{\Lambda}^{-1} \mathbf{z} =  \sum_{i=1}^{n_\text{id}} \frac{z_{\text{id},i}^2}{\lambda_{\text{id},i}} + \sum_{j=1}^{n_\text{ex}} \frac{z_{\text{ex},j}^2}{\lambda_{\text{ex},j}} \, ,
\end{align}
where $\boldsymbol{\Lambda} =
\begin{bmatrix}
\boldsymbol{\Lambda}_{\text{id}} & \mathbf{0} \\
\mathbf{0} & \boldsymbol{\Lambda}_{\text{ex}}
\end{bmatrix}$ includes the eigenvalues $\lambda$ from both identity and expression subspaces. 

We compared the performance of the proposed model with the state-of-the-art baby face model, the BabyFM \cite{Morales2025}. For this purpose, the BabyFM was fitted to the same unseen subjects.

\subsubsection*{Assessing realism and expression control in 3D synthesis}

We evaluated BabyFlow’s ability to generate realistic and expressive 3D infant faces by randomly sampling from the identity and expression latent spaces, resulting in $\mathbf{z}_\text{id}$ and $\mathbf{z}_\text{ex}$ latent codes, under the assumption of a multivariate Gaussian distribution. These latent codes were then mapped to the shape parameter space using the learned identity and expression flows, resulting in $\mathbf{w}_{\text{id}}$ and $\mathbf{w}_{\text{ex}}$, which were then combined through the bilinear model to reconstruct the 3D facial mesh $\mathbf{x}$. We show qualitative examples of the generated 3D shapes to visually assess their realism and diversity.

To further evaluate BabyFlow’s ability to represent unseen expressions and identities, and to examine the structure and continuity of the learned latent spaces, we performed three types of interpolations:
\begin{itemize}
    \item \textbf{Expression interpolation:} keeping the identity fixed, we selected pairs of expression latent codes $\mathbf{z}_{{ex_1}}$ and $\mathbf{z}_{{ex_2}}$ corresponding to different facial expressions and computed intermediate codes via linear interpolation:
\begin{equation}
    \mathbf{z}_{{ex}\nu} = (1 - \nu)\mathbf{z}_{{ex_1}} + \nu \mathbf{z}_{{ex_2}}, \quad \nu \in [0, 1]
\end{equation}
Each interpolated code $\mathbf{z}_{\text{ex}\nu}$ was then decoded into $\mathbf{w}_{\text{ex}\nu}=f^{-1}(\mathbf{z}_{\text{ex}\nu})$ and passed through the bilinear model (together with a fixed identity code $\mathbf{w}_{\text{id}}=f^{-1}(\mathbf{z}_{\text{id}\nu})$) to generate a 3D mesh. 
    \item \textbf{Identity interpolation:} keeping expression fixed, we interpolated between identity latent codes  $\mathbf{z}_{{id_1}}$ and $\mathbf{z}_{{id_2}}$ in the same manner as for expression.
    \item \textbf{Joint interpolation:} both identity and expression codes were interpolated simultaneously as described above.
\end{itemize}


This allowed us to assess whether the latent spaces are continuous and well-structured, producing smooth, artifact-free transitions across expressions, identities, and their combinations.

\subsubsection*{Assessing diffusion-based 2D face synthesis conditioned on BabyFlow}
To evaluate the potential of BabyFlow for cross-modal (geometry-to-image) face synthesis, we explored the generation of realistic 2D infant images conditioned on the 3D infant scans produced by the model. We employ the open-source diffusion-based image generator Stable Diffusion 1.5\footnote{\url{https://github.com/huggingface/diffusers}} (SD 1.5) \cite{Rombach_2022_CVPR}, in combination with a multi-ControlNet framework \cite{controlnet}, to synthesize high-quality and photorealistic 2D facial images. The conditioning mechanism integrates two complementary geometric cues extracted from the 3D mesh: a depth map and a Canny edge map. The depth map is obtained by rendering the 3D face in a random pose view and capturing the z-buffer using the PyTorch3D library\footnote{\url{https://pytorch3d.org/}}, while the edge map is generated by applying a Canny filter to the grayscale version of the same rendered image. These two maps are used as input to separate ControlNet branches, which guide the generation process to align with the 3D geometry provided by the BabyFlow model. To further steer the visual appearance of the generated image, we employ a structured text prompt in the format: \textit{“Realistic frontal view portrait of an/a [ethnicity] infant”}, where the ethnicity token is manually set to either Black, Chinese, Indian, Filipino, White or Hispanic. This cross-modal setup, which combines 3D structural cues with semantic textual guidance, enables the synthesis of 2D infant portraits that are not only photorealistic but also faithful in morphology and expression to a given 3D scan.

\subsubsection*{Assessing the latent representations} 
\label{sec:sampling}
\begin{figure*}[ht]
  \centering
    \includegraphics[width= 1 \textwidth]{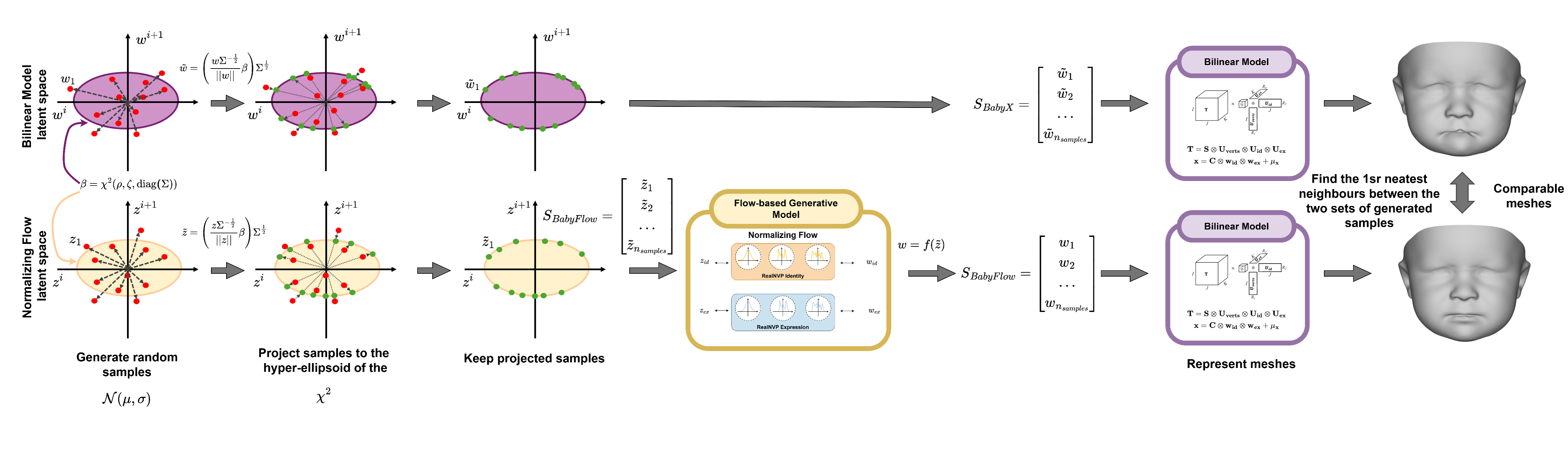}
      \caption{Sampling pipeline to compare meshes generated with and without applying the NFs. First, we generate random samples from a multivariate Gaussian distribution. Then, we project the generated samples to the hyperellipsoid defined by Mahalanobis distance $\beta$. Finally, the projected samples are introduced in the bilinear model to obtain the corresponding 3D shape representation. To ensure comparing similar shapes, we compute the nearest neighbor of the $\mathbf{w}$ parameters.}
      \label{fig:sampling}
\end{figure*} 
A common assumption in the modeling of 3DMMs is that the latent spaces capturing identity and expression variations follow multivariate Gaussian distributions. To assess the validity and implications of this assumption, we conducted an ablation study comparing and analyzing the structure of the latent spaces before and after applying the NFs. 

First, we examined the statistical properties of these spaces using kernel density estimation and visualized them with k-density plots. Second, we analyzed how the assumption of Gaussianity affects the 3D sample generation. Let us denote by $\mathbf{b} \in \mathbb{R}^D$ a latent vector sampled from a multivariate Gaussian distribution. This vector can represent either the latent codes before the NFs transformation (i.e., without-NF: $\mathbf{w}_{\text{id}}$ or $\mathbf{w}_{\text{ex}}$) or after the NFs transformation (i.e., with-NF: $\mathbf{z}_{\text{id}}$ or $\mathbf{z}_{\text{ex}}$), depending on the model being assessed. The samples were randomly generated according to the Gaussian distribution: $\mathbf{b} \sim \mathcal{N}( \boldsymbol{\mu}, \Sigma)$, where $\boldsymbol{\mu}\in \mathbb{R}^D$ is the mean of the distribution and $\Sigma \in \mathbb{R}^{D \times D}$ is the covariance matrix. Since the latent vectors are assumed to be Gaussian distributed, the squared Mahalanobis distance from any point $\mathbf{b}$ to the mean $\boldsymbol{\mu}$ follows a chi-squared distribution. The confidence region at level $\rho$ is represented by the hyperellipsoid whose squared Mahalanobis distance is less than or equal to the critical value $\beta^2=\chi^2_\zeta(\rho)$, where $\chi^2_\zeta(\rho)$ is the upper (100$\rho$)-th percentile of a chi-squared distribution with $\zeta$ degrees of freedom. To fairly compare samples from before and after the NFs transformation, all samples $\mathbf{b}$ were projected onto this hyperellipsoid, ensuring they have the same relative distance from the mean:
\begin{equation}
\tilde{\mathbf{b}} =\frac{ ( \mathbf{b} - \boldsymbol{\mu})\beta }{||(\mathbf{b} -\boldsymbol{\mu})^T \Sigma^{-1/2}||} + \boldsymbol{\mu}   
\end{equation}
where $\boldsymbol{\mu}$ and $\Sigma$ denote the empirical mean and covariance of the latent space, and $\beta^2 = \chi^2_\zeta(\rho)$ is the critical value of the chi-squared distribution with $\zeta$ degrees of freedom. This normalization ensures that all compared samples lie at equal statistical distances from the mean, capturing equivalent variation in each latent space. 

To further evaluate local consistency, we identified nearest neighbors in the without-NF mesh space for each projected with-NF sample. This allowed qualitative comparison of similarly located facial representations across models. The rationale behind this approach is illustrated in Figure~\ref{fig:sampling}, where a 2D visualization of the hyperellipsoid demonstrates how samples are matched.

Finally, we applied parallel analysis \cite{Camras1993} to identify the number of statistically significant components in the bilinear model (i.e., the model before applying the NFs). We found that the expression space contains 6 significant dimensions, while the identity space contains 26. These dimensionalities define the degrees of freedom $\zeta$ used in the chi-squared projections for our comparative analysis. Specifically, for the projections, we used a confidence level $\rho = 0.99$, with $\zeta_{ex} = 7$ and $\beta_{ex} = 4.07$ when sampling expressions and  $\zeta_{id} = 26$ and $\beta_{id} = 6.01$ when sampling identities.



\backmatter

\bmhead{Funding declaration}

This work is partly supported by MICIU/AEI/10.13039/501100011033/ under the project grant PID2024-161292OB-I00 and PRE2021-097544 scholarship, the ICREA Academia programme and the NIH Eunice Kennedy Shriver National Institute of Child Health \& Human Development grant R42 HD08171203.

\bmhead{Ethics Statement}
The data used to construct the BabyFlow model consist of 3D head photography scans of pediatric patients collected at Children’s National Hospital, Washington, DC, USA. These data contain protected health information; therefore, it cannot be made publicly available in compliance with HIPAA regulations. The study was reviewed and approved by the Institutional Review Board of Children’s National Hospital (Washington, DC). All methods were performed in accordance with the relevant guidelines and regulations, including the principles of the Declaration of Helsinki (1964) and its later amendments (2013). Written informed consent was obtained from the parents or legal guardians of all participating minors for the use of their data for research purposes. To protect patient privacy, no patient facial scans are shown in the manuscript. All figures and visualizations presented are synthetic 3D facial scans or 2D images generated by the trained model and do not represent identifiable individuals.

\bmhead{Data Availability}
Due to privacy concerns, the original scan data used in this study cannot be shared publicly, as they contain human-identifiable information. However, the model is built upon the BabyFM framework, and both the resulting newborn model and associated templates are available upon reasonable research request at \url{http://fsukno.atspace.eu/BabyFaceModel.htm}. The FaceWarehouse (FW) dataset is also accessible upon request from its original authors at \url{http://kunzhou.net/zjugaps/facewarehouse/}. The resulting mapping between the FaceWarehouse and BabyFM models, as well as the bilinear model and the resulting NF model, has been deposited on Zenodo under restricted access (DOI:  \url{https://doi.org/10.5281/zenodo.17477552}). Access will be granted to qualified academic researchers for non-commercial use upon reasonable request through the Zenodo platform.

\bmhead{Code Availability}
The code used to develop and analyze the models in this study is available at [ \url{https://doi.org/10.5281/zenodo.17477552}]. The implementation relies on standard Python and MATLAB scientific computing libraries, including TensorLy, and other publicly available packages. Additional scripts for data processing and model construction are deposited alongside the NF model on Zenodo under the same restricted-access conditions.

\noindent


\begin{appendices}

\section{Anatomical Landmarks}\label{secA1}

\begin{figure}[H]
  \centering
    \includegraphics[width= 0.5 \textwidth]{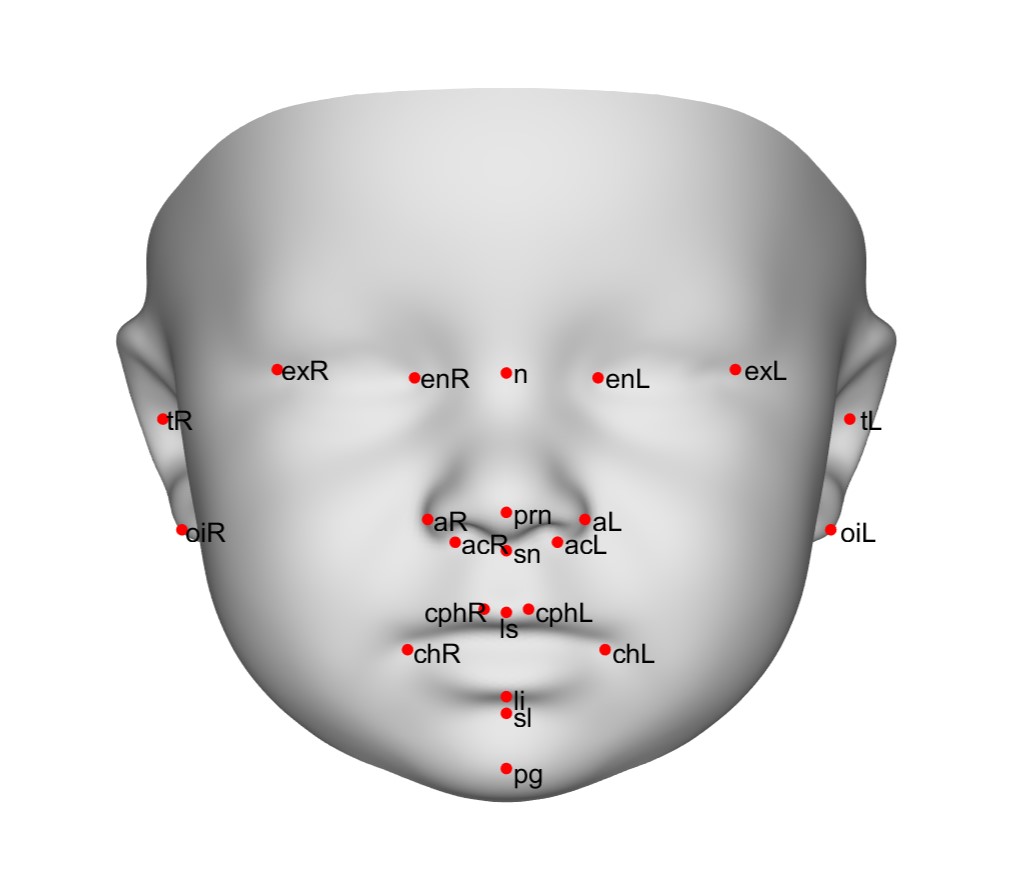}
      \caption{Illustration of the 23 anatomical landmarks considered in the BabyFM \cite{Morales2025}. For facial expressions recognition, we focus on the 19 central landmarks, excluding those located in the ears. Landmark abbreviations: exR = exocanthion Right; enR = endocanthion Right; n = nasion; enL = endocanthion Left; exL = exocanthion Left;  aR = alare Right; acR = alar crest Right; prn = pronasale; aL = alare Left; acL = alar crest Left; sn = subnasale; chR = cheilion Right; cphR = crista philtrum Right;  ls = labiale superius; chL = cheilion Left; cphL = crista philtrum Left; li = labiale inferius; sl = sublabiale; pg = pogonion; tR = tragion Right; oiR = otobasion inferius Right; tL = tragion Left; and oiL = otobasion inferius Left.}
      \label{fig:lmks}
\end{figure} 
\section{Facial expressions}\label{secA1}
\begin{table}[htbp]
	\centering
	\begin{tabular}{|c|c|c|}
	\hline
	\textbf{} & \textbf{Facial Expression} & \textbf{FACS AU} \\
	\hline
	\textbf{1} & Neutral & AU 0  \\
	\hline
	\textbf{2} & Close Eye Left & AU 43 \\
	\hline
	\textbf{3} & Close Eye Right & AU 43  \\
	\hline
	\textbf{4} & Squint Left & AU 44 \\
	\hline
    \textbf{5} & Squint Right & AU 44   \\
	\hline
	\textbf{6} & Slit Left & AU 42   \\
	\hline
	\textbf{7} & Slit Right & AU 42  \\
	\hline
	\textbf{8} & Lid Tightener Left & AU 7   \\
	\hline
	\textbf{9} & Lid Tightener Right & AU 7  \\
	\hline
	\textbf{10} & Lid Droop Left & AU 41  \\
	\hline
	\textbf{11} & Lid Droop Right & AU 41  \\
	\hline
	\textbf{12} & Inner Brow Raiser Left & AU 1\\
	\hline
	\textbf{13} & Inner Brow Raiser Right & AU 1\\
	\hline
    \textbf{14} & Upper Lid Raiser Left & AU 5  \\
	\hline
	\textbf{15} & Upper Lid Raiser Right & AU 5  \\
	\hline
	\textbf{16} & Brow Lower Left & AU 4 \\
	\hline
	\textbf{17} & Brow Lower Right & AU 4   \\
	\hline
	\textbf{18} & Sad Eyes & -  \\
	\hline
	\textbf{19} & Outer Brow Raiser Left & AU 2 \\
	\hline
	\textbf{20} & Outer Brow Raiser Left & AU 2  \\
	\hline
	\textbf{21} & Jaw Thrust & AD 29 \\
	\hline
	\textbf{22} & Jaw Left & -  \\
	\hline
    \textbf{23} & Mouth Stretch & AU 27   \\
	\hline
	\textbf{24} & Lip Slider Left & AD 30   \\
	\hline
	\textbf{25} & Lip Slider Right & AD 30  \\
	\hline
	\textbf{26} & Jaw Right &  -  \\
	\hline
	\textbf{27} & Lip Depressor Left & AU 15  \\
	\hline
	\textbf{28} & Lip Depressor Right & AU 15  \\
	\hline
	\textbf{29} & Lip Corner Puller Left & AU 12 \\
	\hline
	\textbf{30} & Lip Corner Puller Right & AU 12 \\
	\hline
	\textbf{31} & Dimpler Left & AU 14  \\
	\hline
    \textbf{32} & Dimpler Right & AU 14  \\
	\hline
	\textbf{33} & Lip Stretcher Left & Au 20  \\
	\hline
	\textbf{34} &  Lip Stretcher Right & AU 20  \\
	\hline
	\textbf{35} & Lip Suck Left & AU 28  \\
	\hline
	\textbf{36} & Lip Suck Right & AU 28\\
	\hline
	\textbf{37} & Upper Lip Raiser & AU 10  \\
	\hline
	\textbf{38} & Jaw Drop & AU 26 \\
	\hline
	\textbf{39} & Lip part & AU 25  \\
	\hline
	\textbf{40} & Lip Pucker & AU 18  \\
	\hline
	\textbf{41} & Lip Tightener & AU 23  \\
	\hline
    \textbf{42} & Chin Raiser & AU 17   \\
	\hline
	\textbf{43} & Lip Funneler & AU 22   \\
	\hline
	\textbf{44} & Nose Wrinkler & AU 9 \\
	\hline
	\textbf{45} & Cheek Puffer & AU 13  \\
	\hline
	\textbf{46} & Check Raiser Left & AU 6 \\
	\hline
	\textbf{47} & Check Raiser Right & AU 6   \\
    \hline

	\end{tabular}
    \caption{Facial expressions that conform the expression space of the constructed bilinear model. They correspond to the 47 action units (AUs) of the Facial Action Coding System (FACS) available in the FaceWarehouse (FW) dataset \cite{Zhou2014}.}
\end{table}
	
\begin{table}[htbp]
	\centering	
	\begin{tabular}{|c|c|c|}
	\hline
	\textbf{} & \textbf{Facial Expression} & \textbf{AUs of FW}\\
	\hline
	\textbf{48} & Happy &  [29,30,39,46,47] \\
	\hline
	\textbf{49} & Surprise & [10,11,19,20,14,15,23] \\
	\hline
	\textbf{50} & Disgust & [44,37] \\
	\hline
    \textbf{51} & Fear & [10,11,10,20,14,15,33,34,39]  \\
	\hline
	\textbf{52} & Angry & [10,11,16,17,33,34,39]  \\
	\hline
	\textbf{53} & Sad & [16,17,27,28] \\
	\hline
	\textbf{54} & Happily Surprised & [10,11,19,20,29,30,39,14,15,38]  \\
	\hline
	\textbf{55} & Happily Disgusted & [37,29,30,39,44,46,47] \\
	\hline
	\textbf{56} & Sadly Fearful & [10,11,16,17,39,33,34]  \\
	\hline
	\textbf{57} & Sadly Surprised & [10,11,16,17,39,38]  \\
	\hline
	\textbf{58} & Sadly Angry & [16,17,27,28,42] \\
	\hline
	\textbf{59} & Sadly Disgusted & [16,17,37,10,11,46,47,42] \\
	\hline
	\textbf{60} & Fearfully Angry & [16,17,33,34,39]  \\
	\hline
	\textbf{61} & Fearfully Surprised  & [10,11,19,20,14,15,33,34,39,38]  \\
	\hline
	\textbf{62} & Fearfully Disgusted   & [10,11,16,17,37,33,34,39,19,20]  \\
	\hline
	\textbf{63} & Disgustedly Surprised & [10,11,19,20,14,15,37,42]  \\
	\hline
	\textbf{64} & Appalled  & [16,17,37,42] \\
	\hline
	\textbf{65} & Hatred & [16,17,37] \\
	\hline
	\textbf{66} & Awed  & [10,11,19,20,14,15,39,38,33,34]   \\
	\hline
	\textbf{67} & Angrily Disgusted  & [16,17,37,42,44,8,9] \\
	\hline
	\textbf{68} & Angrily Surprised & [16,17,39,38,8,9]  \\
	\hline
	\textbf{69} & Lip Corner Puller with closed eyes & [2,3,23,29,30]  \\
	\hline
	\textbf{70} & Closed Eyes & [2,3] \\
	\hline
	\textbf{71} & Open mouth with closed eyes &  [2,3,23] \\
	\hline
	\textbf{72} & Lip Sunk with closed eyes &  [2,3,35,36] \\
	\hline
	\textbf{73} & Lip Punker with closed eyes & [2,3,40] \\
	\hline
	\textbf{74} & Chin Raiser with closed eyes & [2,3,42] \\
	\hline
	\textbf{75} & Lip Fuller with closed eyes & [2,3,43] \\
	\hline
	\textbf{76} & Crying with closed eyes & [2,3,44,46,47,39,23,33,34] \\
	\hline
	
	\end{tabular}
	
	\caption{Composed facial expressions that define the expression space of the constructed bilinear model. Each expression is created as a combination of action units (AUs). The table specifies the AU combinations in the FaceWarehous (FW) \cite{Zhou2014} dataset used to generate the corresponding facial expressions.}
\end{table}




\end{appendices}


\bibliography{sn-bibliography}

\end{document}